\documentclass[lettersize,journal]{IEEEtran}
\IEEEoverridecommandlockouts

\usepackage{cite}
\usepackage{amsmath,amssymb,amsfonts}
\usepackage{algorithmic}
\usepackage{graphicx}
\usepackage{amsfonts}
\usepackage{amssymb}
\usepackage{textcomp}
\usepackage[T1]{fontenc}
\usepackage{longtable} 
\usepackage{supertabular} 
\usepackage{xcolor}
\usepackage{subcaption}
\usepackage{tabularx}
\usepackage{comment}
\usepackage{booktabs}
\usepackage{mathrsfs} 
\usepackage{float} 
\usepackage{booktabs}

\usepackage{algorithmic}\makeatletter
\usepackage{algorithm}  
\usepackage{multirow}
\usepackage{makecell}
\usepackage{hyperref}

\allowdisplaybreaks
\def\BibTeX{{\rm B\kern-.05em{\sc i\kern-.025em b}\kern-.08em
    T\kern-.1667em\lower.7ex\hbox{E}\kern-.125emX}}
\begin{document}

\title{Interaction-Aware Trajectory Prediction for Safe Motion Planning in Autonomous Driving: A Transformer-Transfer Learning Approach
 \\

\author{Jinhao Liang, \textit{Member, IEEE}, Chaopeng Tan, Longhao Yan, Jingyuan Zhou, Guodong Yin, \textit{Senior Member, IEEE}, Kaidi Yang, \textit{Member, IEEE}%
\thanks{This research was supported by the Singapore Ministry of Education (MOE) its Academic Research Fund Tier 1 (A-8001183-00-00). This article
solely reflects the opinions and conclusions of its authors and not the Singapore MOE or any other entity. (Corresponding author: Kaidi Yang).

Jinhao Liang, Kaidi Yang, Longhao Yan and Jingyuan Zhou are with the Department of Civil and Environmental Engineering, National University of Singapore, Singapore, 119077, (E-mail: \{jh.liang, kaidi.yang\}@nus.edu.sg, \{longhao.yan, jingyuanzhou\}@u.nus.edu).

Chaopeng Tan is with the Department of Transport and Planning, Delft University of Technology, Gebouw 23, Stevinweg 1, 2628 CN, Delft, Netherlands (email: c.tan-2@tudelft.nl).

Guodong Yin is with the School of Mechanical Engineering, Southeast University, Nanjing, 211189, China. (E-mail: ygd@seu.edu.cn).
}
\thanks{This work has been submitted to the IEEE for possible publication. Copyright may be transferred without notice, after which this version may no longer be accessible.}

}}
\maketitle

\begin{abstract}
A critical aspect of safe and efficient motion planning for autonomous vehicles (AVs) is to handle the complex and uncertain behavior of surrounding human-driven vehicles (HDVs).  Despite intensive research on driver behavior prediction, existing approaches typically overlook the interactions between AVs and HDVs assuming that HDV trajectories are not affected by AV actions. To address this gap, we present a transformer-transfer learning-based interaction-aware trajectory predictor for safe motion planning of autonomous driving, focusing on a vehicle-to-vehicle (V2V) interaction scenario consisting of an AV and an HDV. Specifically, we construct a transformer-based interaction-aware trajectory predictor using widely available datasets of HDV trajectory data and further transfer the learned predictor using a small set of AV-HDV interaction data. Then, to better incorporate the proposed trajectory predictor into the motion planning module of AVs, we introduce an uncertainty quantification method to characterize the predictor’s errors, which are integrated into the path-planning process.
Our experimental results demonstrate the value of explicitly considering interactions and handling uncertainties. 

\end{abstract}


\section{Introduction}
\IEEEPARstart{A}{utonomous} driving technology is pivotal in revolutionizing transportation safety and efficiency, offering unprecedented potential to reduce accidents and congestion while improving accessibility for all travelers \cite{petrovic2020traffic, liang2022mas, liang2023polytopic}. 
However, as the penetration rates of autonomous vehicles (AVs) can only increase incrementally due to the gradual enhancement of technological maturity and social acceptance, AVs are envisioned to operate in mixed traffic environments alongside human-driven vehicles (HDVs) \cite{yue2023cooperative, sun2020cooperative, zhou2024enhancing, zhou2024parameter}.  
One challenge associated with mixed-traffic environments is the motion planning of AVs, which involves the planning of AV trajectories considering the interactions with surrounding HDVs \cite{shawky2020factors}.  
Since human driver behavior can be highly complex and stochastic, the key to addressing this challenge is (i) to 
 predict HDV behavior accurately and (ii) to optimize AV trajectories based on the noisy predictions of HDV’s future trajectories.

Driver behavior prediction is crucial for the deployment of AVs in mixed-traffic environments, as it enables AVs to anticipate and react to the actions of HDVs \cite{singh2021analyzing}. With the continuous advancement of sensor technologies and machine learning algorithms, researchers have made significant progress in modeling and predicting diverse driver behaviors, such as lane changing, braking, and turning \cite{mozaffari2020deep}. These prediction models often leverage a combination of real-time sensor data, history driving data, and contextual information to anticipate driver intentions and actions \cite{xu2022aggressive,liao2023driver, wang2021intelligent}. 
Early methods utilize physics-based models \cite{lefevre2014survey} or conventional shallow learning algorithms \cite{gao2021trajectory, liu2020early} such as Support Vector Machines, Dynamic Bayesian Networks, and Hidden Markov Models. However, these models are not able to account for the variability and unpredictability of human driving behavior. In recent years, advancements in deep learning \cite{girma2020deep, phillips2017generalizable} have introduced more intricate models to capture complex driver behaviors in realistic driving conditions. 

Despite intensive research on driver behavior prediction, existing approaches typically ignore the \emph{interactions} between AVs and HDVs and rely on simplified assumptions that the predicted HDV trajectories are independent of the future actions of AVs \cite{mcnaughton2011motion, schwarting2018planning}.   
In other words, existing approaches lack the capacity to incorporate ``what-if'' analyses, assuming fixed reactions from surrounding vehicles regardless of the ego AV's actions. Nevertheless, interactions among vehicles frequently shape one another's behaviors. Thus, \emph{interaction-aware} trajectory prediction is crucial for the safe and efficient navigation of AVs in mixed-traffic environments. 
A small portion of existing works have investigated interaction-aware trajectory prediction by developing deep learning-based algorithms. For example, reference \cite{alahi2016social} employs a social pooling structure to understand the spatial interactions among pedestrians, utilizing hidden states produced by an LSTM network, which is further refined by reference \cite{deo2018convolutional} via convolutional layers. Recent studies \cite{song2020pip, chen2022efficient, espinoza2022deep} present a novel Planning-Informed Trajectory Prediction (PiP) framework, which incorporates the ego vehicle's future planning information into the prediction process. The comparative test results also demonstrate that the PiP framework can better improve the accuracy of the trajectory predictor in driving scenarios involving complex interactions between vehicles. To enhance the interaction capability both between vehicles and between vehicles and the road, many state-of-the-art prediction methods utilize transformer framework \cite{ngiam2021scene, zhang2022trajectory} to achieve high prediction accuracy.


However, to the best of our knowledge, existing interaction-aware deep learning-based trajectory prediction algorithms suffer from two limitations. First, existing deep learning methods \cite{sadeghian2019sophie, gupta2018social, kingma2013auto, lee2017desire, ngiam2021scene, zhang2022trajectory} train the trajectory predictor using only HDVs’ trajectory data. This is due to the limited availability of interactive trajectory data between AVs and HDVs in existing trajectory databases, such as NGSIM, HighD, and Waymo. Nevertheless, as the response of HDVs to AVs may be fundamentally different from their response to other HDVs \cite{wen2023analysis}, the developed methods may not accurately capture the interactions between AVs and HVs, thereby reducing the trajectory prediction performance in mixed traffic. Second, recent interaction-aware trajectory prediction algorithms \cite{song2020pip, chen2022efficient, espinoza2022deep} do not incorporate the uncertainty of prediction errors into path planning. Neglecting trajectory predictor uncertainty in an AV's path planning process can lead to overlooking possible deviations in other vehicles' behavior, thereby increasing the risk of collisions. 

Hence, in this work, we aim to use limited trajectory data from real-world autonomous driving scenarios to further enhance the adaptability of the trajectory predictor in mixed traffic scenarios. Furthermore, we also use the PiP framework in \cite{song2020pip} to integrate the HDV’s trajectory predictor into the path-planning process. However, different from studies in \cite{song2020pip, chen2022efficient, espinoza2022deep}, we further integrate the uncertainty of trajectory prediction results into the path planning. Specifically, an uncertainty quantification method is introduced to feature potential prediction errors and is incorporated into path planning as a safety constraint for obstacle avoidance. Moreover, a trajectory-tracking controller has been developed to regulate the AV's motion as it navigates along the reference path.

\emph{Statement of Contribution}. The main contributions of this work are two-fold. First, unlike existing trajectory predictors \cite{sadeghian2019sophie, gupta2018social, kingma2013auto, lee2017desire, ngiam2021scene, zhang2022trajectory} that are trained using only HDV trajectory data, we propose a novel transformer-transfer learning-based trajectory predictor that leverages a small amount of interaction data between AVs and HDVs from real-world scenarios to further enhance the adaptability of the predictor to mixed-traffic scenarios. Specifically, we construct a transformer-based interaction-aware trajectory predictor using a large dataset of HDVs’ trajectory data and further transfer the learned predictor using a small set of AV-HDV interaction data to account for mixed-traffic scenarios. 
Second, we introduce an uncertainty quantification method to characterize the predictor’s errors, which are integrated into the path-planning process. By ensuring safety constraints, the trajectory selector would balance various performance indicators to generate a reference trajectory for the tracking controller.

The remainder of this paper is organized as follows. Section \uppercase\expandafter{\romannumeral2} presents the methodological framework. Sections \uppercase\expandafter{\romannumeral3}, and \uppercase\expandafter{\romannumeral4} present the designs of the trajectory predictor, path-planning algorithm, and trajectory-tracking controller, respectively. Section \uppercase\expandafter{\romannumeral5} performs case studies and analyzes simulation results, and Section \uppercase\expandafter{\romannumeral6} concludes the paper. 

\section{Problem Description and Systematic Framework} 
\subsection{Problem Description}

\begin{figure}[ht]
	\centering
	\includegraphics[width=3.1 in]{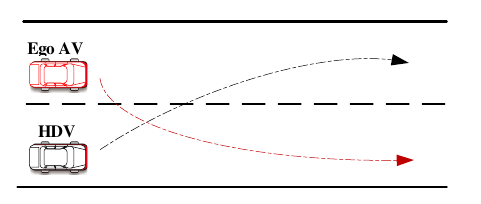}
	\caption{V2V encounter scenario.} \label{Fig1:V2V encounter scenario}
\end{figure}

As illustrated in Fig. \ref{Fig1:V2V encounter scenario}, we consider a vehicle-to-vehicle (V2V) interaction scenario where an AV and an HDV seek to perform lane changes. Such a scenario is common in the early deployment stages of AVs with limited penetration rates. We treat the AV as the ego vehicle and the HDV as the encountering vehicle. The goal is to safely and efficiently control the AV, considering the potential interactions with the HDV. 

\begin{figure}[ht]
	\centering
	\includegraphics[width=2.8in]{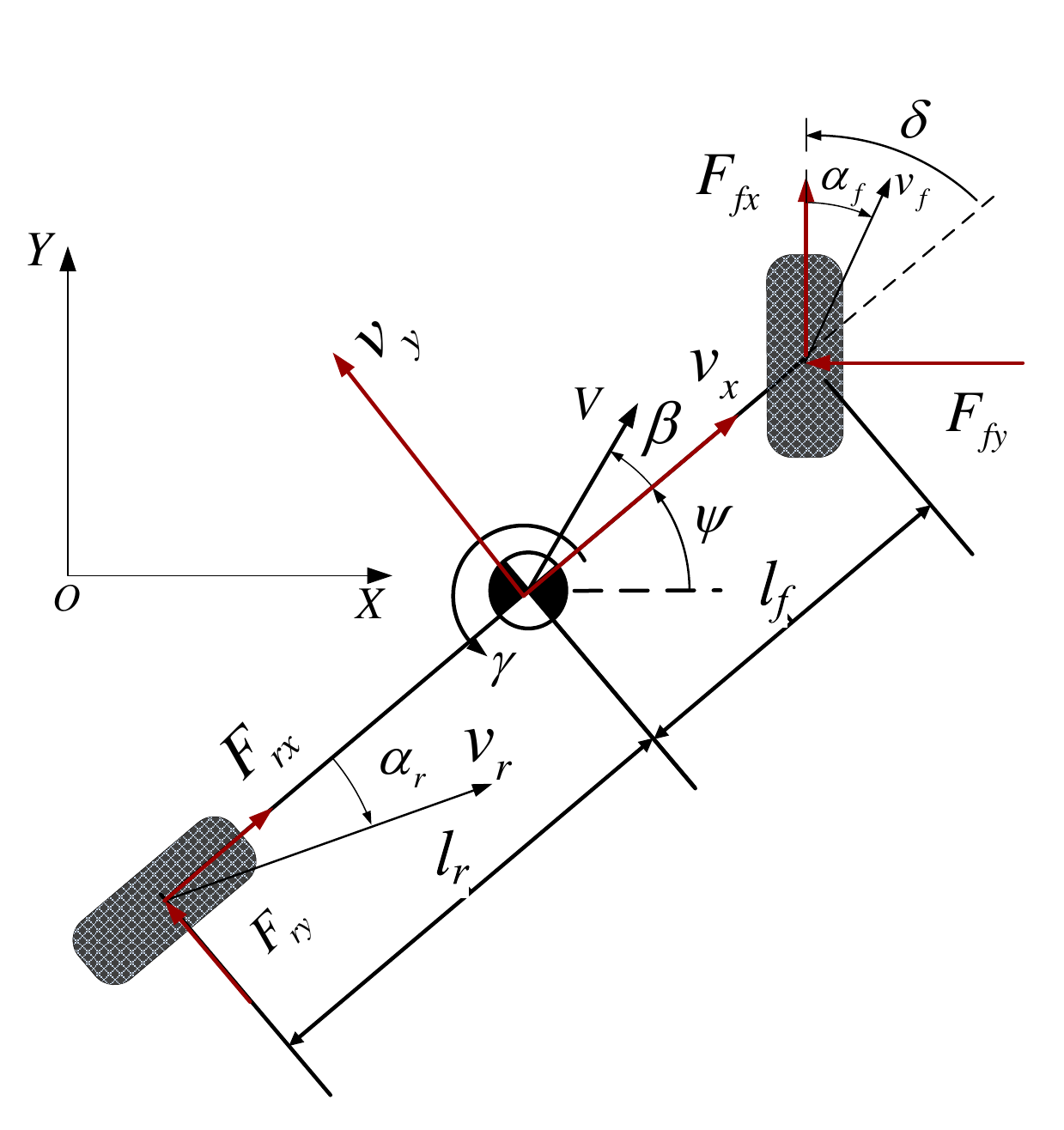}
	\caption{Vehicle dynamics model.}\label{fig1:vehicle dynamics model}
\end{figure}

The dynamics of the AV are characterized by a widely used bicycle model \cite{liang2021distributed} demonstrated in Figure \ref{fig1:vehicle dynamics model} with model parameters provided in Table \ref{tab1:vehicle model parameters}. 
Specifically, the vehicle dynamics model can be represented by a state-space equation as follows.
\begin{align} \label{eq1:state-space equation}
\dot{\xi}=A \xi+B u,
\end{align}
with
\begin{align*}
A\!&=\!\begin{bmatrix}
    0 & 0 & 1 & 0 & -v_{y} & 0 \\ 0 & 0 & 0 & 1 & v_{x} & 0\\
    0 & 0 & 0 & 0 & 0 & v_{y} \\ 0 & 0 & 0 & \frac{2\left(c_{r}+c_{f}\right)}{m v_{x}} & 0 & \frac{2\left(c_{f} c_{f}-c_{r} l_{r}\right)}{m v_{x}}-v_{x} \\ 0 & 0 & 0 & 0 & 0 & 1 \\ 0 & 0 & 0 & \frac{2\left(c_{f} l_{f}-c_{r} l_{r}\right)}{J_{z} v_{x}} & 0 & \frac{2\left(c_{r} l_{r}^{2}+c_{f} l_{f}^{2}\right)}{J_{z} v_{x}} \end{bmatrix}, \\    
B\!&=\!\left[\begin{array}{cccccc} 0 & 0 & 1 & 0 & 0 & 0 \\ 0 & 0 & 0 & \frac{-2 c_{f}}{m} & 0 & \frac{-2 c_{f} l_{f}}{J_{z}}\end{array}\right]^{T},
\end{align*}
where the vehicle state $\xi=[x,y,v_x,v_y,\psi,\gamma]^T$ includes the AV's global longitudinal position $x$, global lateral position $y$, longitudinal velocity $v_x$, lateral velocity $v_y$, yaw angle $\psi$, and yaw rate $\gamma$. The control input of the AV is represented by $u=[a_x,\delta]^T$, where $a_x$ and $\delta$ indicate the longitudinal acceleration and steering input, respectively. Variables $m$ and $J_z$ are the vehicle mass and the inertia moment of yaw direction, respectively, $c_f$ and $c_r$ denote the cornering stiffness of the front and rear tires, respectively, and $l_f$ and $l_r$ indicate the distances from the vehicle center of gravity (CoG) to the front and rear axles, respectively.

\begin{table}[ht]
	\begin{center}
		\caption{Vehicle Configuration Parameters}
		\begin{tabular}{ c  c  c}
			\hline
			\toprule  
			Symbol & Description & Value[units] \\
			\midrule  
$J_{z}$ & Inertia yaw moment of the vehicle & $606.1~\left(\mathrm{kg} \cdot \mathrm{m}^{2}\right)$ \\   
			$m$ & Vehicle mass & $1274~(\mathrm{kg})$ \\
$c_{f}$ & Cornering stiffness of front tire & $85000~(\mathrm{N} / \mathrm{rad})$ \\
$c_{r}$ & Cornering stiffness of rear tire & $112000(\mathrm{~N} / \mathrm{rad})$ \\
$l_{f}$ & Distance from CoG to front axle & $1.016~(\mathrm{m})$ \\
$l_{r}$ & Distance from CoG to rear axle & $1.562~(\mathrm{m})$ \\
			\bottomrule 
			\hline 
		\end{tabular}
  \label{tab1:vehicle model parameters}
	\end{center}
\end{table}
 
\begin{figure*}[t]
	\centering
	\includegraphics[width=6.2 in]{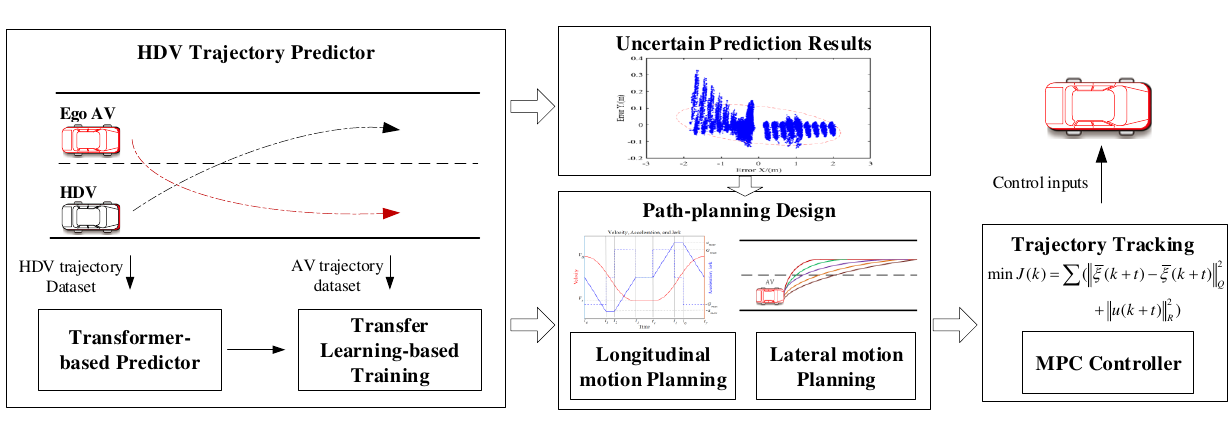}
	\caption{The Systematic framework.} \label{fig2:The Systematic framework}
\end{figure*}
 
\subsection{Methodological Framework}
Our methodological framework is illustrated in Fig. \ref{fig2:The Systematic framework}, which involves an HDV prediction module (Section III) and an AV path-planning module (Section IV). 

The HDV prediction module aims to predict the future trajectories of the HDV, which serves as a crucial input to the AV motion planning module. Here, the main challenge is that the behavior of the HDV highly depends on the decisions made by the AV, which requires the trajectory prediction algorithm to be interaction-aware. Moreover, the HDV response to the AV may be different from its response to another HDV. Therefore, the HDV prediction model trained fully based on HDV naturalistic driving data, as constructed in most existing studies, may not be suitable to characterize the HDV's interactions with the AV. 
To address these challenges, we first train an interaction-aware transformer-based predictor based on a real-world HDV driving dataset and then leverage a transfer learning framework to derive the HDV model that better characterizes HDV's interactions with AVs, using limited autonomous driving data.   

The AV path-planning module would generate the candidate paths and then select the optimal reference trajectory from these paths. Since the AV path-planning module takes the HDV trajectory prediction as input, its performance significantly relies on the accuracy of the prediction. To address this challenge, we explicitly quantify the uncertainties of the prediction module by evaluating the prediction module's performance on online data. Such uncertainties are explicitly incorporated into the motion planning of AVs to derive the reference trajectory, which is tracked via a model predictive controller (MPC). 


\begin{figure*}[t]
	\centering
	\includegraphics[width=6.2 in]{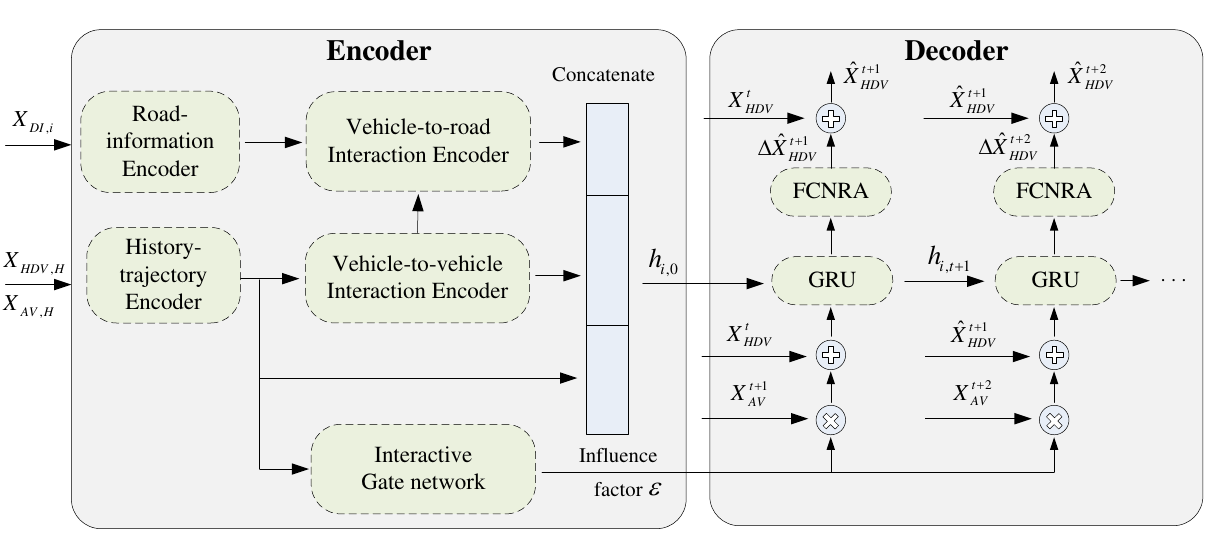}
	\caption{Interaction-aware based predictor.} \label{fig3:Interaction-aware based Predictor}
\end{figure*}

\section{A Transformer-Transfer Learning-based Trajectory Predictor for HDVs}
This section introduces an interaction-aware HDV trajectory prediction model that considers the potential response of HDV trajectories to AV decisions. Section~\ref{sec_MDP} presents a transformer-based interaction-aware HDV trajectory prediction model, and Section~\ref{sec_transfer} introduces a transfer learning approach for training the proposed trajectory prediction model to adapt to the limited availability of the interaction data between HDVs and AVs. Section III-C presents a quantification method to describe the uncertain prediction results.


 \subsection{Interaction-Aware Trajectory Prediction Model} \label{sec_MDP}
In this section, we propose an interaction-aware transformer-based model for predicting the future trajectories of HDVs, which captures the response of HDVs to AV decisions by including the planned trajectories of AVs as input to the HDV trajectory prediction model. 
Mathematically, let $X_{\text{AV}}^t$ and $X_{\text{HDV}}^t$ represent the positions of the AV and the HDV at time step $t$, respectively. 
Then, we use the following information to predict the HDV's future trajectory $\hat{X}_{\text{HDV,P}}^t=\{\hat{X}_{\text{HDV}}^{t'}\}_{t'=t+1}^{t+t_p}$: (i) the history trajectory of the AV, denoted by $X_{\text{AV,H}}^t=\{X_{\text{AV}}^{t'}\}_{t'=t-t_h+1}^{t}$, and the history trajectory of the HDV, denoted by $X_{\text{HDV,H}}^t=\{X_{\text{HDV}}^{t'}\}_{t'=t-t_h+1}^{t}$, (ii) road information, and (iii) the AV's planned trajectory $X_{\text{AV,P}}^t=\{X_{\text{AV}}^{t'}\}_{t'=t+1}^{t+t_p}$, where $t_h$ and $t_p$ represents the time horizons of history data and predictions, respectively. 



The detailed architecture of the proposed trajectory predictor model is illustrated in Fig. \ref{fig3:Interaction-aware based Predictor}, which includes an encoder component and a decoder component. As shown in Fig.~\ref{fig3:Interaction-aware based Predictor}, the encoder component consists of a history trajectory encoder, a road information encoder, and an interaction encoder between vehicles and the road. 

\vspace{0.5em} 
\noindent \emph{1) History Trajectory and Road Information Encoder}
\vspace{0.5em} 

The history trajectory encoder uses a simple transformer-based architecture consisting of a positional encoding module, a self-attention module, and a feed-forward neural network module. First, a positional encoding module is added to the history trajectories of AVs ($X_{\text{AV,H}}$) and HDVs ($X_{\text{HDV,H}}$) to construct the input sequence. Then, the self-attention module is used to extract deep connections within the input sequence across different historical moments. Finally, we utilize a feed-forward neural network to obtain the feature representation, with the output of the last time step serving as the final feature representation to design the interaction encoder.

The road information encoder consists of two fully connected layers with ReLU activation functions to extract road features. In actual driving conditions, we are most concerned about the road information of the road center, road boundaries, and speed limit. Therefore, we represent the road information in the form of multi-sets of points along the path, with each path point having its unique features $X_{\text{DI},i}=(X_{\text{cen},i},X_{\text{left},i},X_{\text{righgt},i},V_{\text{lim},i})$, where $X_{\text{cen},i}$ and $V_{\text{lim},i}$ represent the road center position and speed limit information at the sampling point $i$, respectively. $X_{\text{left},i}$ and $X_{\text{right},i}$ represent the information of the left and right road boundaries, respectively.

\begin{figure*}[ht]
	\centering
	\includegraphics[width=6.2 in]{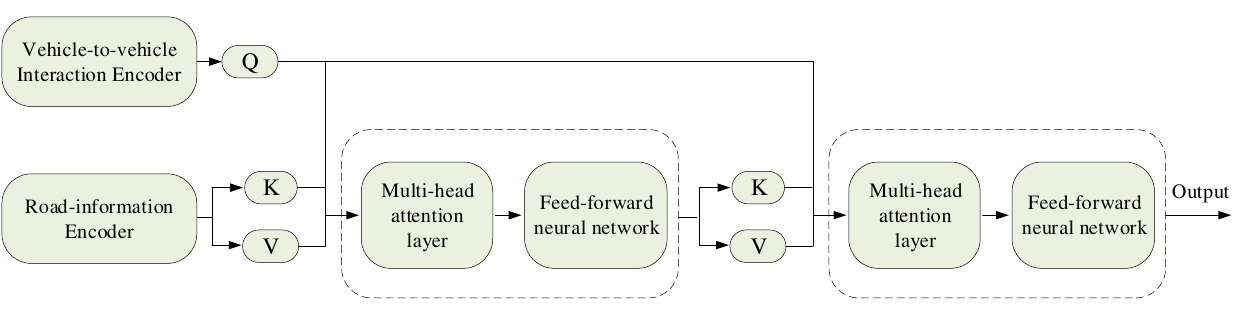}
	\caption{Vehicle-road interaction encoder.} \label{fig4: Vehicle-road interaction encoder}
\end{figure*}

\begin{figure*}[t]
	\centering
	\includegraphics[width=6.2 in]{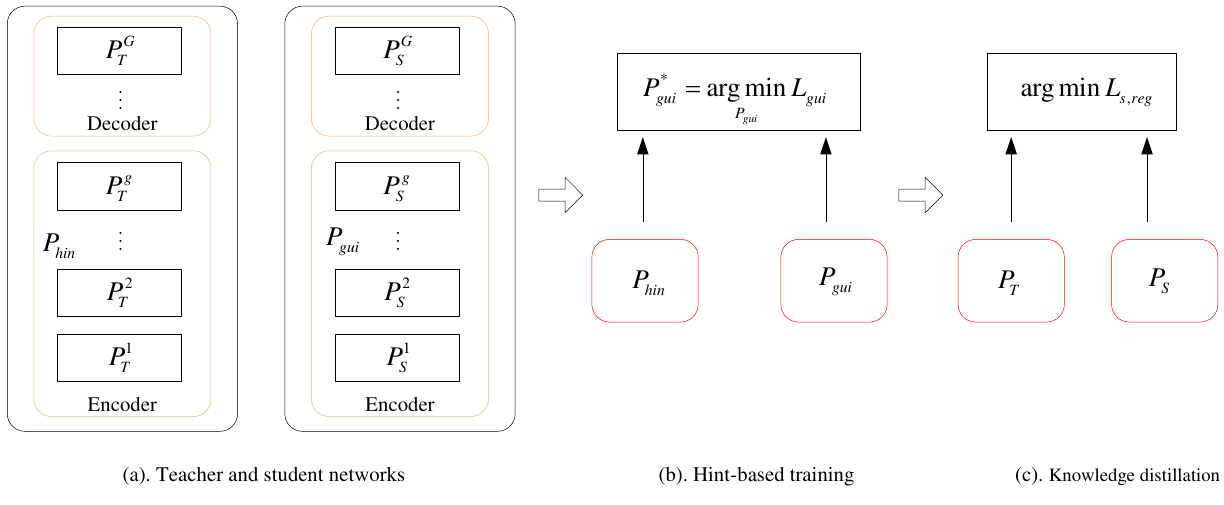}
	\caption{The training process of the transfer-learning framework.} \label{fig5:The transfer-learning framework}
\end{figure*}

\vspace{0.5em} 
\noindent \emph{2) Interaction Encoder}
\vspace{0.5em}

Interaction behavior is modeled using both a vehicle-to-vehicle interaction encoder and a vehicle-to-road interaction encoder. The key to modeling these interactions in trajectory prediction lies in capturing the mutual influences and dynamic dependencies among agents to accurately predict their future behaviors. The transformer’s attention mechanism enhances this process by enabling the model to focus on relevant interactions, thereby improving the prediction of complex behaviors through a better understanding of how each agent influences others in dynamic environments \cite{wang2023lane}.

Specifically, we adopt a two-layer self-attention module to extract vehicle-to-vehicle interaction features. This is because when faced with complex interactions between vehicles, the dual self-attention modules not only increase the model's depth but also enhance its adaptability to different scenarios \cite{chen2022vehicle}. The two-layer self-attention modules are connected using a residual connection architecture, which allows for the integration of the original input features into the second layer and helps the model more accurately extract and understand the relationships between vehicles in complex interaction scenarios \cite{veit2016residual}.

Similar to the vehicle-to-vehicle interaction module, the vehicle-to-road interaction encoder (see Fig. \ref{fig4: Vehicle-road interaction encoder}) includes two cross-attention modules to extract interaction information. These cross-attention modules can effectively combine information from different sources (such as vehicle-to-vehicle interactions and vehicle-to-road environment interactions), capturing important relationships and thereby improving prediction accuracy. Each cross-attention module consists of a multi-head attention layer and a feed-forward neural network. The vehicle-to-vehicle encoder output, serving as a query, enables the model to focus on dynamic interactions between vehicles, while the feature vectors from the road-information encoder, used as both key and value, enhance the model's understanding of road information and environmental constraints. Then, two-layer cross-attention modules are employed to extract the interaction features between driving agents and roads.

Finally, the output from the history-trajectory encoder, vehicle-to-vehicle encoder, and vehicle-to-road encoder are concatenated into $h_{i0}$ and fed into the decoder to predict the HDVs' future trajectories.

\vspace{0.5em} 
\noindent \emph{3) Decoder Design}
\vspace{0.5em} 

We next present the decoder, as shown in Fig. \ref{fig3:Interaction-aware based Predictor}, which converts the output of the encoder to the predicted future HDV trajectories. Specifically, the main component of the decoder is $t_p$ gated recurrent units (GRUs), each corresponding to a future time step $t'=t+1, \cdots, t+t_p$. 
These GRUs are chained through the hidden states, which are initialized by the concatenated output from the encoder component (denoted by $h_i^{0}$). 
The input to the GRU corresponding to time step $t'$ is designed to be a weighted combination of the HDV state at time step $t'-1$ and the planned AV state at time step $t'$, i.e.,   
\begin{align}  \label{eq2: the decoder output}
X_{\text{GRU,in}}^{t'}=\varepsilon X_{\text{AV}}^{t'}+\hat{X}_{\text{HDV}}^{t'-1}
\end{align}
where we set $\hat{X}_{\text{HDV}}^{t} = X_{\text{HDV}}^{t}$ for $t'=t+1$ since the current HDV state can be accurately measured. It is worth noting that the future planned trajectory of the AV is introduced into the GRU to enable interaction awareness among HDVs. The weight parameter $\epsilon$ ranging between 0 and 1 is derived from an interactive gate network in the encoder component, which builds on a fully connected network with the output mapped to the interval $[0,1]$ via a sigmoid function. In complex traffic environments, predicting future trajectories involves uncertainty. The introduction of the weight parameter $\varepsilon$ allows the model to adjust its evaluation of different trajectory possibilities based on the current environment and history behavior, thereby better addressing this uncertainty \cite{huang2022survey}.



The hidden state of each GRU is further processed through a single-layer fully connected network with a ReLU activation function (FCNRA) to obtain the predicted change in HDV states $\Delta{\hat{X}_{\text{HDV}}^{t'}}$, which is further accumulated to calculate the predicted HDV states $\hat{X}_{\text{HDV}}^{t'},~t'=t+1,\cdots,t+t_p$. 

\subsection{Transfer Learning-Based Training Approach } \label{sec_transfer}
One challenge for training the interaction-aware trajectory prediction model presented in Section \ref{sec_MDP} is the limited availability of the interaction data between HDVs and AVs. 
To address this challenge, we propose a transfer learning framework that adapts the model trained on HDV trajectory data to capture the interactions between AVs and HDVs. Specifically, we first use a large amount of HDVs' trajectory data to train the prediction model, which serves as a teacher network. Then, we leverage a transfer learning framework and a small amount of interaction data between HDVs and AVs to adapt the teacher network to a student network with a similar architecture. The reason for adopting the transfer learning framework is based on the observation that the HDV-AV interactions and HDV-HDV interactions share similarities in the fundamental reasoning processes of human drivers. 
Consequently, the transfer learning framework can exploit knowledge learned about HDV interactions to accelerate the training of the interaction-aware HDV prediction model in mixed-traffic environments, despite limited autonomous driving data.


Due to the modular design of the trajectory predictor, which includes an encoder and a decoder, it becomes feasible to incorporate a hint-based training process \cite{bae2019layer, adriana2015fitnets} within the traditional knowledge distillation framework \cite{hinton2015distilling}. In this modular transfer-learning process, we utilize not only the final output of the whole teacher network but also the intermediate representations learned by the encoder as hints to enhance the training effectiveness of each module and improve the overall performance of the predictor \cite{adriana2015fitnets}.


\vspace{0.5em} 
\noindent \emph{1) Classical Knowledge Distillation Process}
\vspace{0.5em} 

Knowledge distillation is a machine learning technique where a large, complex model (the teacher) transfers its knowledge to a smaller, simpler model (the student) \cite{hinton2015distilling}. Hence, learning a good regression model from the teacher network is important to ensure prediction accuracy of the student network. However, the teacher's regression guidance may sometimes contradict the true reference direction \cite{chen2017learning}. Therefore, the output of the teacher's regression model is used as an upper limit for the student model to achieve, rather than as a reference value in this work. Typically, we expect the student's regression model to align closely with the true reference value. Once the performance of the student network behaves better than that of the teacher by a certain margin, we will not impose additional loss on the student model. It is called the bounded regression loss $L^T$ of the teacher model, which serves as the foundation for constructing the regression loss $L_{\text{reg}}^S$ of the student model.


\begin{align}  \label{eq3: the bounded regression loss of the teacher}
&L^T(\hat{X}_{\text{HDV,P}}^T, \hat{X}_{\text{HDV,P}}^S, X_{\text{HDV}}^R)=\\
&\left\{\begin{aligned}
&||\hat{X}_{\text{HDV,P}}^S-X_{\text{HDV}}^R||_2^2, ~ &\ \text{if} \ \ ||\hat{X}_{\text{HDV,P}}^S-X_{\text{HDV}}^R||_2^2+n \nonumber 
\\&&>||\hat{X}_{\text{HDV,P}}^T-X_{\text{HDV}}^R||_2^2 \\
 &~ 0,\qquad \text{otherwise}
\end{aligned}\right. \\
&L_{\text{reg}}^S=L_{\text{smo}}(\hat{X}_{\text{HDV,P}}^S,X_{\text{HDV}}^R)+\nu L^T(\hat{X}_{\text{HDV,P}}^T,\hat{X}_{\text{HDV,P}}^S, X_{\text{HDV}}^R) \label{eq4:the regression loss of the student model} 
\end{align}
where $\hat{X}_{\text{HDV,P}}^T$ and $\hat{X}_{\text{HDV,P}}^S$ represent the prediction values of the teacher network and student network, respectively. $X_{\text{HDV}}^R$ is the true reference value. $L^S_{\text{reg}}$ is the constructed regression loss for the student network used for the training. $n$ indicates the boundary value, and $v$ is a scale factor (set as 0.4 in this work). From Eqs. \eqref{eq3: the bounded regression loss of the teacher}-\eqref{eq4:the regression loss of the student model}, it is clear that when the prediction error of the student network exceeds that of the teacher network, the regression loss $L_{\text{reg}}^S$ will include the loss of the teacher network. Furthermore, $L_{\text{smo}}$ represents the smooth $L_s$ loss. i.e., ($||\hat{X}_{\text{HDV,P}}^S-X_{\text{HDV}}^R||_2^2$), which aims to improve the training accuracy \cite{girshick2015fast}, where

\begin{align}  \label{eq5: smooth loss}
&L_{smoL_s}=\\
&\left\{\begin{aligned}
&0.5*||\hat{X}_{\text{HDV,P}}^S-X_{\text{HDV}}^R||_2^2,\qquad 
if |\hat{X}_{\text{HDV,P}}^S-X_{\text{HDV}}^R|<1\\ \nonumber
&|\hat{X}_{\text{HDV,P}}^S-X_{\text{HDV}}^R|-0.5, \qquad otherwise
\end{aligned}\right.
\end{align}

Our integrated loss motivates the student network to approximate or exceed the teacher network's performance in regression, without excessively urging the student once parity with the teacher is achieved.

\vspace{0.5em} 
\noindent \emph{2) Hint-Based Training Process}
\vspace{0.5em} 

The classical knowledge distillation process only employs the final output of the network. Reference \cite{bae2019layer} illustrates that leveraging the intermediate outputs (i.e., hints) provided by the teacher network can improve the training process and enhance the performance of the student network. Hints provide intermediate features and gradient information, improving gradient flow and information transmission during training, thereby enhancing the depth and effectiveness of networks. Additionally, leveraging hints allows for more efficient use of existing data, thereby reducing the need for large-scale datasets \cite{adriana2015fitnets}. Inspired by this, we introduce a hint-based training process to improve network performance. Specifically, we utilize the hidden layer's output of the teacher network (which is called the hint layer) to guide the training of the student network (which is called the guided layer). In this work, we choose the encoder components as the hint and the guided layers for the teacher and student networks, respectively. Then we train the network of the guided layer using the following loss function.
\begin{align}  \label{eq6: Loss of the hint}
L_{gui}=\frac{1}{2}||N_{hin}(P_{hin})-N_{gui}(P_{gui})||_2^2
\end{align}
where $N_{hin}$ and $N_{gui}$ denote the nested functions of the hint and guided layers' networks, respectively, characterized by network parameters $P_{hin}$ and $P_{gui}$. Specifically, the nested functions in this work are constructed using the concatenated features $h_{i0}$ of the encoder.

\vspace{0.5em} 
\noindent \emph{3) The Combined Training Process}
\vspace{0.5em}

We combine the aforementioned training methods, i.e., classical knowledge distillation and hint-based training, to construct the transfer learning framework. An overview of the transfer learning-based training process is provided in Fig. \ref{fig5:The transfer-learning framework}, where $P_T$ and $P_S$ indicate all network parameters of the teacher and student networks, respectively. First, we randomly initialize the student network parameters $P_S$ and train the teacher network parameters $P_T$ using a large amount of HDV's trajectory data ( Fig. \ref{fig5:The transfer-learning framework} (a)). Then, we extract the encoder from the trained teacher network for hint-based training and obtain the guiding layer parameters $P_{gui}$ for the student network using limited interaction data between HDVs and AVs (Fig. \ref{fig5:The transfer-learning framework} (b)). Finally, we employ the knowledge distillation method to train the whole network using the pre-trained parameters of the guided layer (Fig. \ref{fig5:The transfer-learning framework} (c)). Algorithm 1 provides a detailed outline of the training process.

\begin{algorithm}[ht]\label{tab1:The algorithm flow of the decision-making RL}
	\caption{The algorithm flow of the transfer learning framework}
	\renewcommand{\algorithmicrequire}{\textbf{Input:}}
	\renewcommand{\algorithmicensure}{\textbf{Initialize:}}
	\begin{algorithmic}[1]
		\REQUIRE The trained network parameters $P_{T}$, and $P_{hint}$.
        \ENSURE $P_{S}$, and $P_{gui}$
        \STATE $P_{hin}\leftarrow(P_{hin}^1,P_{hin}^2,...,P_{hin}^g)$
        \STATE $P_{gui}\leftarrow(P_{gui}^1,P_{gui}^2,...,P_{gui}^g)$
        \STATE $P_{gui}^*\leftarrow{\mathop{argmin}\limits_{P_{gui}}L_{gui}}$
         \STATE Update the network parameters $P_{gui}$ for the student model as $P_{gui}^*$       
         \STATE Perform $\arg\min L_{s,reg}$
         \STATE \textbf{Output:} $P_{S}^{*}$
	\end{algorithmic}
\end{algorithm}

\subsection{Uncertainty Quantification of Predictor}

Although the proposed trajectory predictor can improve HDV trajectory prediction accuracy in mixed traffic by explicitly accounting for the interactions between AVs and HDVs, trajectory prediction errors are still inevitable, which will further deteriorate the path planning performance of AVs and even compromise the safety of AVs. To mitigate the impact of the prediction errors on AV path planning, we explicitly quantify such errors and incorporate the uncertainty quantification results into the AV path planning described in Section \ref{sec:path planning}. Specifically, this work introduces an error ellipse to describe prediction errors in both the $x$ and $y$ directions.


First, an error set is obtained by testing the trained network on the validation data. As the AV's longitudinal and lateral motions are highly correlated, we utilize the variance $\vartheta_x$ and $\vartheta_y$ and the covariance $\vartheta_{x-y}$ to construct a matrix $\Theta$ as follows.

\begin{equation}
\Theta = 
\begin{bmatrix} \vartheta_x^2 & \vartheta_{x-y}  \\ 
                \vartheta_{x-y} & \vartheta_y^2 \end{bmatrix},\nonumber      
\end{equation}

Then, through the eigenvalue transformation, the error ellipse is determined as follows to quantify the uncertainty of the predictor.
\begin{align} \label{eq7:error elipse}
\frac{x^2}{\eta_1}+\frac{y^2}{\eta_2}=\digamma
\end{align}%
where $\eta_1$ and $\eta_2$ represent the largest and smallest values of the eigenvalues for the matrix $\Theta$, respectively. $\digamma$ indicates the chi-square value, which is detailed in reference \cite{lancaster2005chi}. In this work, the chi-square value is selected to ensure a 95$\%$ confidence level, taking into account the search efficiency of the error set, whereby a higher confidence level implies a greater coverage of the error set within the error ellipse.
Furthermore, the rotation angle of the ellipse can be further calculated by:
\begin{align} \label{eq8: rotation angle}
\psi=\arctan\frac{\tau_1}{\tau_2}
\end{align} %
\begin{figure}[ht]
	\centering
	\includegraphics[width=3.3 in]{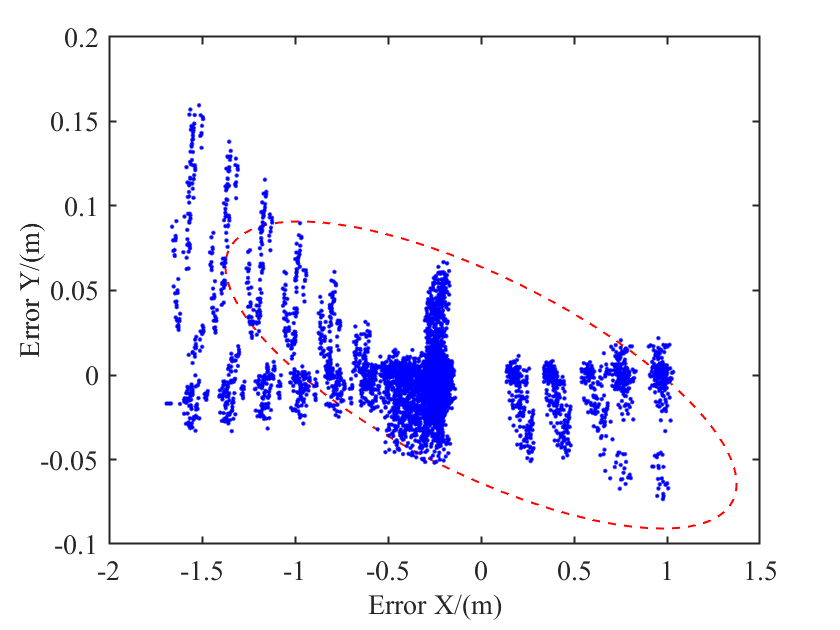}
	\caption{The uncertain representation of the predictor on the Waymo Open Dataset.} \label{Fig6:The uncertain representation of the predictor on the Waymo Open Dataset}
\end{figure}%
where $\tau_1$ and $\tau_2$ represent the basic value and the other element of the maximum eigenvector, respectively. To visually present the uncertain prediction results, we utilize the Waymo Open Dataset to train the predictor and then construct the error ellipse shown in Fig. \ref{Fig6:The uncertain representation of the predictor on the Waymo Open Dataset}. Considering the uncertain prediction results, the constructed ellipse error would be utilized as a constraint to ensure the safety of AVs when generating candidate paths in Section IV-C.

\section{Path Planning and Trajectory Tracking Modules Design} \label{sec:path planning}

We present a path-planning method for generating candidate paths in this section, which will be incorporated into the trajectory predictor in Section III. When selecting the optimal path, we incorporate the uncertainty of trajectory predictions as a safety constraint to enhance the AV's obstacle avoidance capability. After generating an optimal reference trajectory, a trajectory tracking controller is designed using the MPC-based control method.

\subsection{Lateral and Longitudinal Trajectory Planning}
\vspace{0.5em}  

\begin{figure}[ht]
	\centering
	\includegraphics[width=3in]{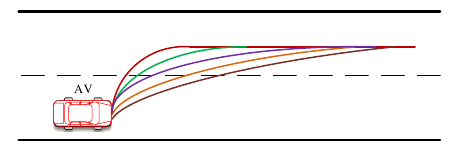}
	\caption{The lateral motion planning.} \label{Fig7:The lateral motion planning}
\end{figure}

\begin{figure}[ht]
	\centering
	\includegraphics[width=3.3 in]{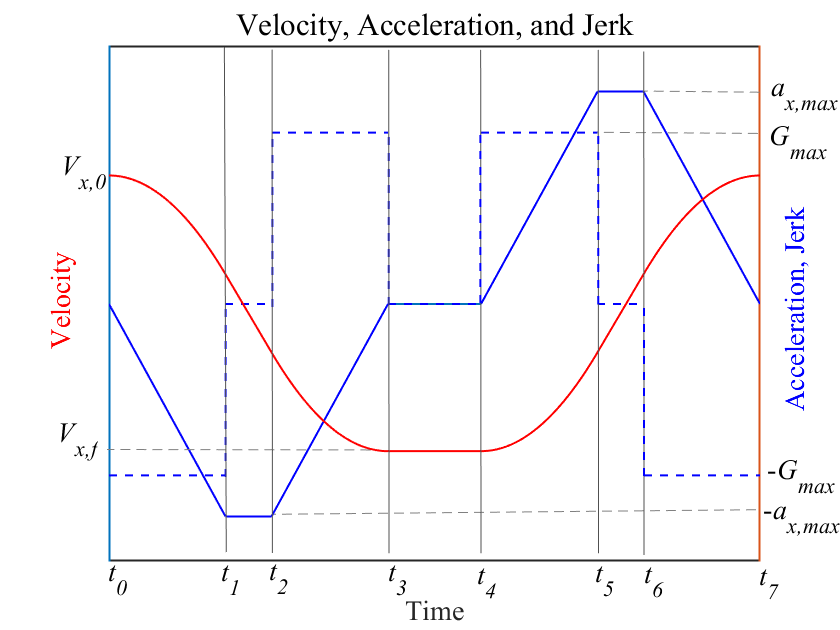}
	\caption{The longitudinal motion planning.}
  \label{fig8:The longitudinal motion planning}
\end{figure}

The traditional trajectory generation method includes both the longitudinal and lateral motion \cite{huang2021personalized} of the AV. In this work, a widely used polynomial representation \cite{li2023dynamic} as shown in Fig. \ref{Fig7:The lateral motion planning} is employed to determine the lateral position, which can be described as follows.
\begin{align} \label{eq11:The reference lateral position}
y_{\text {ref }}=\frac{10 H}{Q^{3}} x_{\text {ref }}^{3}-\frac{15 H}{Q^{4}} x_{\text {ref }}^{4}+\frac{6 H}{Q^{5}} x_{\text {ref }}^{5}
\end{align} %
where $Q$ and $H$ indicate the longitudinal and lateral lane-change distances, respectively. 
The longitudinal lane-change distance $x_\text {ref }$ depends on the AV's velocity planning. In a typical lane-change scenario, the human-driven vehicles would decelerate first to maintain lateral stability and then accelerate to maintain the initial speed control intention. Hence, a trapezoidal acceleration profile \cite{martinez2017assessment} is employed in this work to determine the AV's longitudinal position. Fig. \ref{fig8:The longitudinal motion planning} depicts a velocity profile that integrates acceleration and jerk. where $a_{x,max}$ and $G_{max}$ represent the maximum values of the acceleration and jerk, respectively, while $v_{x,0}$ and $v_{x,f}$ denote the maximum and minimum velocities, respectively. The reference longitudinal position can be calculated as follows:
\begin{align} 
x_{ref,t}&=x_{ref,t-1}+v_{x,t-1}\Lambda{t}+\frac{1}{2}a_{x,t-1}\Lambda_{t}^2+\frac{1}{6}G_{t}\Lambda{t}^3,
\label{eq14:Longitudinal position} 
\end{align}
where $a_{x,t}$, $v_{x,t}$, $x_{ref,t}$ and $G_{t}$ represent the AV's longitudinal acceleration, longitudinal velocity, longitudinal position, and jerk at time $t$, respectively, and $\Lambda{t}$ indicates the time interval.

\subsection{The Optimal Trajectory Selection} 
\vspace{0.5em}

From the Waymo Open Dataset, approximately $\ 85\%$ of lane-change maneuvers occur within the range of 2.5 to 6.5s. Hence, we choose a constant lane-change time with an average value of 4.5s for the AV. Subsequently, candidate paths are generated by configuring different longitudinal speed profiles. The vehicle state on the candidate paths should satisfy the following constraint on vehicle dynamics~\cite{liang2022mas}.
\begin{align} \label{eq23: The dynamics constraint of beta}
\beta_{\text {m}} \leq\left|\tan ^{-1}(0.02 \mu g)\right| 
\end{align}
where $\beta_{m}$ indicates the AV's side slip angle at the sampling point $m$ on the planned path, and $\mu$ is the tire-road friction coefficient.

Furthermore, a safe distance constraint between the AV and HDV should also be satisfied on the candidate paths. However, the HDV trajectory obtained through the trajectory predictor is not accurate. To ensure the safety of the planned path, we incorporate the error ellipse constructed in Section III-C, in addition to the safe distance threshold $S_{\text{saf}}$. This means that on the candidate paths, the distance $S$ between the AV's planned trajectory and the HDV's predicted trajectory should be no lower than the sum of the axis distance $L_{\text{ell}}$ of the error ellipse constructed in Section III-C and the safety threshold $S_{\text{saf}}$.
\begin{align} \label{eq24: Safety distance}
S \geq\left(L_{\text{ell}}+S_{\text{saf}}\right) 
\end{align} 

After generating a set of candidate paths, the Technique for Order Preference by Similarity to Ideal Solution (TOPSIS) method \cite{huang2021personalized} is employed to determine the optimal reference trajectory. We first need to define the following indicators to evaluate the quality of the candidate paths.

\vspace{0.5em} 
\noindent \emph{1) Vehicle safety}
\vspace{0.5em} 

The vehicle safety is quantified by considering the potential collision with the HDVs, which is represented as follows:
\begin{align} 
O_{saf}=\sum_{m=1}^{\epsilon}\exp \Big[\zeta_{1}\left(x_{\text{ref},m}-\widetilde{x}_{\text{HDV},m}\right)^{2}+ \nonumber \\\quad\quad\quad\quad\zeta_{2}\left(y_{\text{ref},m}-\widetilde{y}_{\text{HDV},m}\right)^{2}\Big]  
\label{eq26:The potential collision} 
\end{align}
where $\epsilon$ is the total number of sampling points. $x_{\text{ref},m}$ and $y_{\text{ref},m}$ indicate the AV's longitudinal and lateral positions at the sampling point $m$. Note that the HDV's longitudinal and lateral positions represented by ${x}_{\text{HDV},m}$ and ${y}_{\text{HDV},m}$ are obtained through the transformer-transfer learning-based predictor. 

Considering uncertain prediction results of the HDVs' trajectory, we utilize the modified variables $\widetilde{x}_{\text{HDV},m}$ and $\widetilde{y}_{\text{HDV},m}$ to represent the HDV's longitudinal and lateral positions, respectively. where the point ($\widetilde{x}_{\text{HDV},m}$, $\widetilde{y}_{\text{HDV},m}$) is located on the error ellipse, which is the nearest point to the AV's reference trajectory ($x_{\text{ref},m}$, $y_{\text{ref},m}$). Parameters $\zeta_{1}$ and $\zeta_{2}$ are weight coefficients utilized to standardize the impact of longitudinal and lateral space between the AV and the HDV. Specific configurations can refer to the work \cite{liang2022mas}.

\vspace{0.5em} 
\noindent \emph{2) Vehicle stability}
\vspace{0.5em} 

The vehicle stability performance is evaluated by introducing the side slip angle $\beta$ \cite{liang2021distributed}, which is represented as follows:
\begin{align} 
O_{sta}=\sum_{m=1}^{\epsilon} \beta_{m}^{2}
\label{eq27: Stability performance} 
\end{align}

\vspace{0.5em} 
\noindent \emph{3) Driving comfort}
\vspace{0.5em}

The driving comfort is quantified by the change rate of heading angle $\Delta{\theta}_{m}$ and expressed as follows.
\begin{align} 
O_{com}=\sum_{m=1}^{\epsilon} \Delta{\theta}_{m}^{2}
\label{eq28: Driving comfort} 
\end{align}

After defining the performance indicators for each candidate path, the TOPSIS strategy aims to select the optimal path by selecting the one that is closest to the ideal solution \cite{huang2021personalized}.

\subsection{Trajectory Tracking Controller Design} 
\vspace{0.5em}

After generating an optimal reference trajectory, a trajectory tracking controller is designed using the MPC-based control method. The MPC controller aims to track the reference by constructing a quadratic programming (QP) optimization problem as follows.
\begin{subequations}
    \begin{align} 
\min \quad & J(k)= \sum_{t=1}^{N_{p}}\Big(\left\|\overline{\xi}(k+t \mid k)-\overline{\xi}_{ref}(k+t \mid k)\right\|_Q^{2}\notag \\ &+ 
\left\|u_(k+t \mid k)\right\|_R^{2}\Big) \label{eq36: The objective function}\\
\rm{s.t.}\quad & a_{x, \min } \leq a_{x}(k) \leq a_{x, \max } \label{eq:con1}\\
& \delta_{\min } \leq \delta(k) \leq \delta_{\max } \label{eq:con2} \\
& |\gamma| \leq \frac{\mu g}{V_{x}} \label{eq:con3} \\
& \left|\alpha_{f}\right| \leq \arctan \frac{\mu m g l_{r}}{2 c_{f}\left(l_{f}+l_{r}\right)} \label{eq:con4}
\\
&\left|\alpha_{r}\right| \leq \arctan \frac{\mu m g l_{f}}{2 c_{r}\left(l_{f}+l_{r}\right)} \label{eq:con5}
\end{align}
\end{subequations}
where $\overline{\xi}=[v_x,x,y]^T$, and $\overline{\xi}_{ref}=[v_{x,ref},x_{ref},y_{ref}]^T$. $v_{x,ref}$ represents the reference longitudinal velocity along the reference trajectory. $N_{P}$ is the prediction horizon. The optimization objective function Eq. \eqref{eq36: The objective function} aims to minimize the cost from two aspects: (i) the trajectory-tracking error, and (ii) the control efforts, weighted by the positive-definite matrices $Q\in\mathbb{R}^{3\times3}$ and $R\in\mathbb{R}^{2\times2}$. Eqs. \eqref{eq:con1}-\eqref{eq:con2} impose constraints on the control inputs and the AV's handling stability \cite{liang2022mas}.

Furthermore, the weight coefficients $Q=\operatorname{diag}(Q_1,Q_2,Q_3)$ and $R=\operatorname{diag}(R_1,R_2)$ would impact the tracking performance. In this work, we use a search method to determine the specific value \cite{liang2023energy}. By comparing test results with different weights, $Q_2=Q_3=10Q_1=300$ and $R_1=5R_2=0.5$ are selected to offer a better trade-off among multi-objectives.

\section{Simulation Experiments} \label{Simulation Tests}
This section conducts simulation experiments to validate the proposed framework. The experiment results demonstrate the effectiveness of both the prediction module and the path planning and trajectory tracking module. 

 \subsection{Scenarios and Datasets}
\begin{figure}[ht]
	\centering
	\includegraphics[width=3.5in]{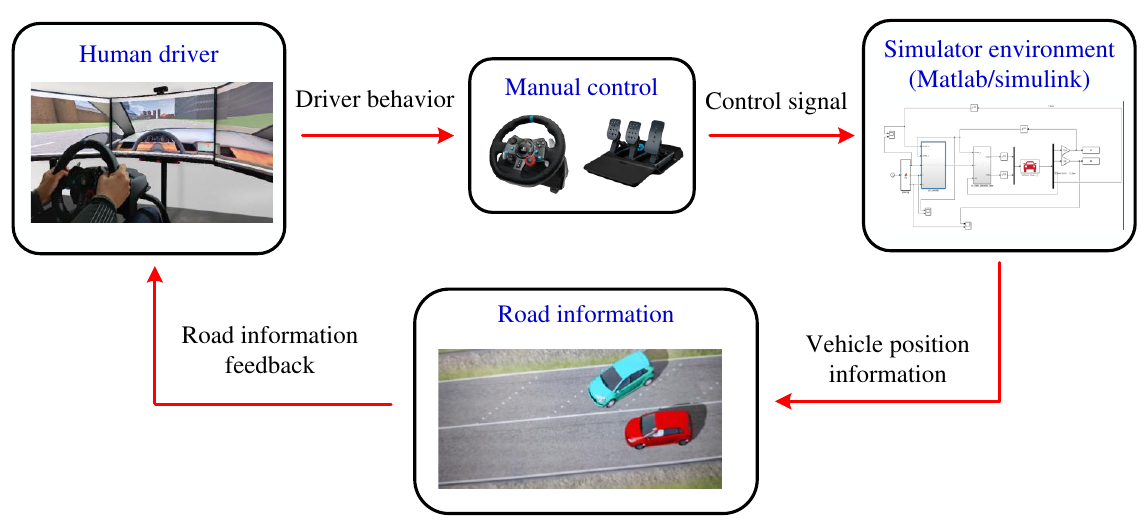}
	\caption{The Simulator Setting.}
 \label{fig 15:The Simulator Setting}
\end{figure}

 We perform the experiments in two scenarios: (i) the V2V interaction scenarios in the Waymo Open Dataset and (ii) a V2V interaction scenario in the driver simulator, where an AV operates along the planned path designed in Section IV, while a human driver performs the lane-change maneuver (see Fig. \ref{fig 15:The Simulator Setting}). For the driver simulator scenario, the commercial software CarSim is used to create a virtual simulation environment with an HDV and an AV, where the HDV is controlled by a driver using an external steering wheel and accelerator pedal, and the AV is controlled with our proposed framework implemented using Matlab/Simulink. 

The reason for considering these two scenarios is two-fold. First, we aim to show that our proposed interaction-aware trajectory prediction algorithm can be used in different scenarios with various human driving behavior and AV control algorithms. Second, the driver simulator experiment is used to validate the driving performance of integrating the proposed interaction-aware trajectory prediction algorithm and our proposed path-planning and motion control algorithms. Note that the interactions between HDVs and AVs rely on the control algorithms of AVs, i.e., the HDV behavior may change if the AV controller changes. Therefore, we cannot use the Waymo Open Dataset for evaluating the integration of our proposed interaction-aware trajectory prediction and control algorithms, as Waymo data is generated by different and unknown AV control algorithms. 

 \subsection{Performance of the Transformer-Transfer Learning-based Predictor}

In this subsection, we evaluate the performance of the transformer-transfer learning-based interaction-aware trajectory predictor. 
To assess the predictive accuracy, we employ two indicators for trajectory prediction: the minimum Average Distance Error (ADE) and the minimum Final Distance Error (FDE). The ADE calculates the average deviation of each sampling point in the predicted trajectories from the true reference, whereas the FDE quantifies the deviation between the final point of the predicted trajectories and the true reference. The prediction errors for two indicators in this work are calculated across all validation data.

\begin{figure}[ht]
	\centering
 \begin{subfigure}{0.48\textwidth}
 \centering
	\includegraphics[width=3.4in]{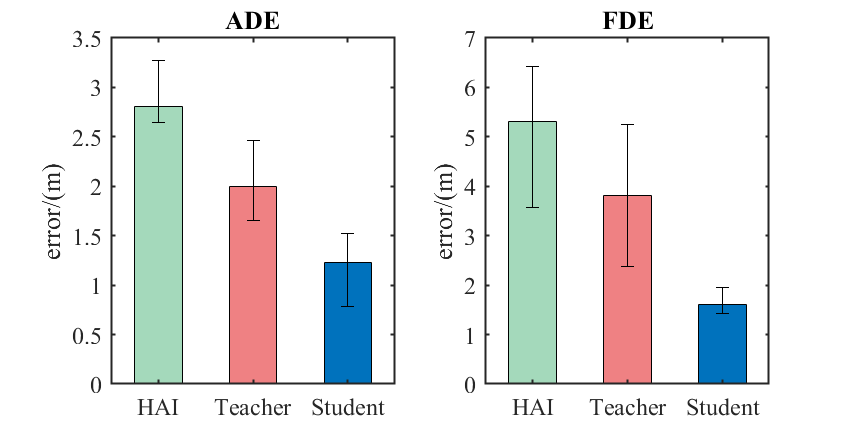}
	\caption{Waymo Open Dataset.}
 \label{fig 10:Comparison of trajectory prediction accuracy with different networks on the Waymo Open Dataset}
 \end{subfigure}
	\begin{subfigure}{0.48\textwidth}
	\centering
	\includegraphics[width=3.4in]{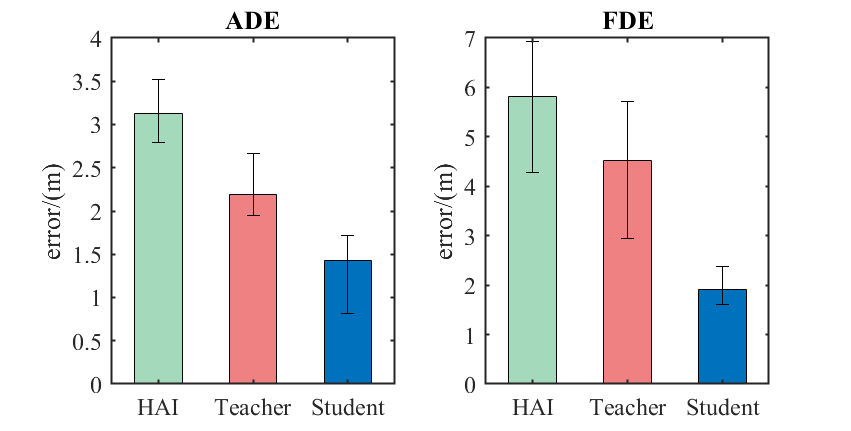}
	\caption{Driver Simulator Dataset.}
 \label{fig 16:Comparison of trajectory prediction accuracy with different networks on the simulator test data}
  \end{subfigure}
\caption{Comparison of trajectory prediction accuracy with different networks.}
 \label{fig: different networks}
\end{figure}%


To facilitate the training of the proposed predictor on both datasets, we normalize the trajectory data by setting the longitudinal position of the AV as the origin and marking the lateral position based on the lanes. Then the trajectory data are divided into two parts: (i) separate HDVs' interaction scenarios involving lane-exchange maneuvers, and (ii) HDV-AV interaction scenarios involving lane-exchange maneuvers. The former data is used to train the trajectory predictor and serves as the teacher network. Then, the teacher network would be employed to guide the learning process of the student network using the latter data. For the training of the teacher network, we use $\ 80\%$ of the trajectory data to train the network, and the remaining data are used as the validation data. In terms of the training process of the student network, due to the limited autonomous driving scenarios, $\ 90\%$ of the trajectory data is employed to train the network, while the remaining data are used as validation data.

\begin{figure}[ht]
	\centering
 	\begin{subfigure}{0.48\textwidth}
	\centering
	\includegraphics[width=3.4in]{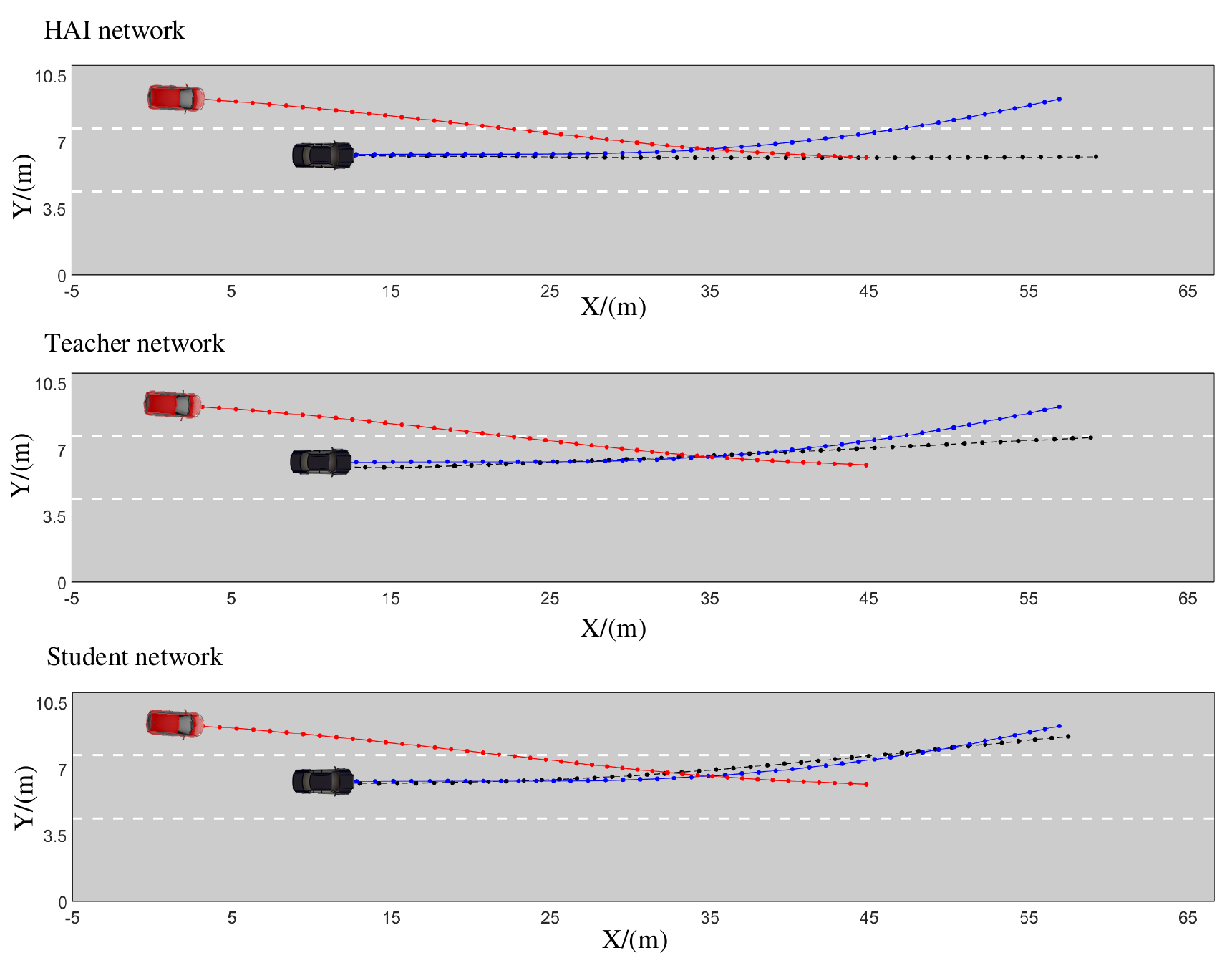}
	\caption{Waymo Open Dataset.}
 \label{fig 12: The trajectory prediction results on the Waymo Open Dataset in a low-speed driving scenario}
 \end{subfigure}
	\begin{subfigure}{0.48\textwidth}
	\centering
	\includegraphics[width=3.4in]{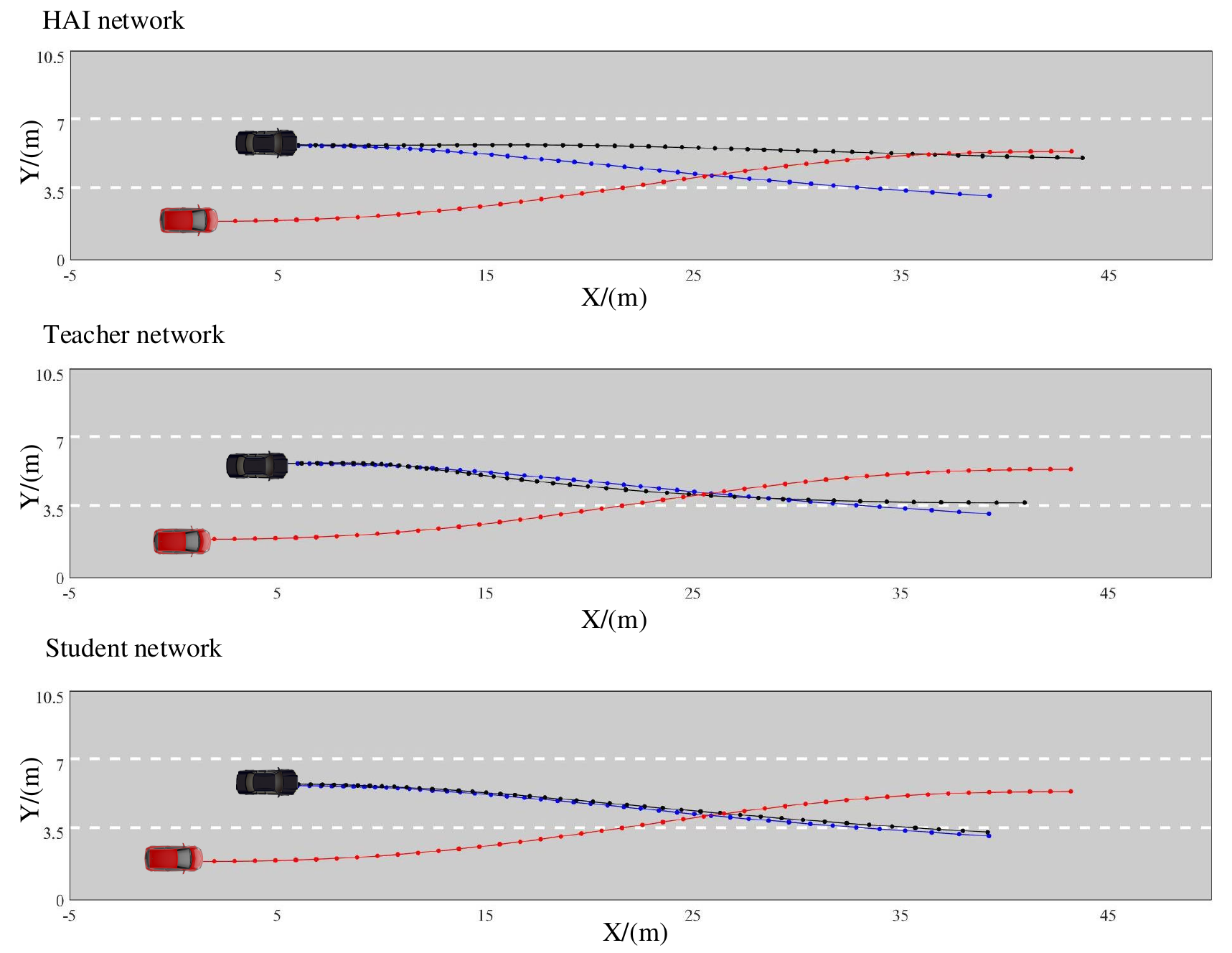}
	\caption{Driver Simulator Dataset.}
 \label{fig 18: The trajectory prediction results on the simulator test data in a low-speed driving scenario}
  \end{subfigure}
  	\caption{The trajectory prediction results in a low-speed driving scenario.}
\label{fig: prediction low-speed driving scenario}
\end{figure}%

\vspace{0.2em}\noindent \textbf{Value of using transfer learning method}. 
We show the value of using transfer learning method by comparing three benchmarks: (i) the trained teacher network (i.e., without transfer learning), (ii) the trained student network (with transfer learning), and (iii) a trained network using only the trajectory data of the HDV-AV interaction scenarios (HAI network). 
The test results are shown in Fig. \ref{fig 10:Comparison of trajectory prediction accuracy with different networks on the Waymo Open Dataset} and Fig. \ref{fig 16:Comparison of trajectory prediction accuracy with different networks on the simulator test data} for the Waymo dataset and the driver simulator dataset. From these results, it is clear that the student network (i.e., the proposed training framework) exhibits the highest predictive accuracy among the three methods. For the Waymo dataset, the indicator of average ADE  can be reduced by $\ 56.26\%$ and $\ 37.42\%$, while the indicator of average FDE can be reduced by $\ 69.60\%$ and $\ 56.86\%$ compared to the HAI network and teacher network, respectively. For the driver simulator dataset, the average ADE shows reductions of $\ 54.33\%$ and $\ 34.97\%$, while the average FDE demonstrates reductions of $\ 70.02\%$ and $\ 57.45\%$, compared to the HAI network and teacher network, respectively. This demonstrates that the transformer-transfer learning-based framework effectively improves predictor accuracy in AV-HDV interaction scenarios despite limited training data. In addition, since we can collect more trajectory data from HDVs than AVs, the prediction results of the trained teacher network perform better than those of the HAI network. 

\begin{figure}[ht]
	\centering
 \begin{subfigure}{0.48\textwidth}
     \includegraphics[width=3.4in]{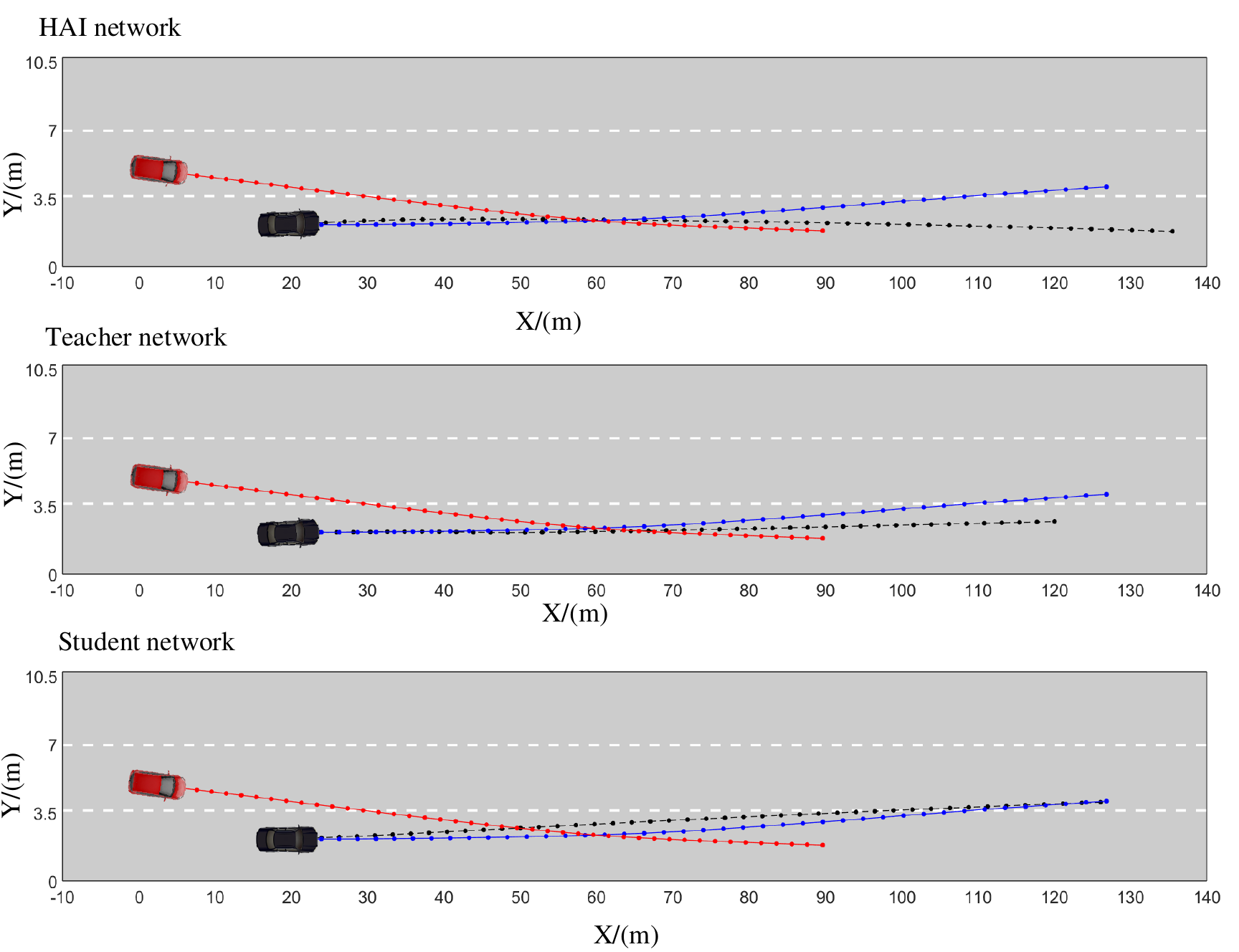}
	\caption{Waymo Open Dataset.}
 \label{fig 13: The trajectory prediction results on the Waymo Open Dataset in a high-speed driving scenario}
 \end{subfigure}
	\begin{subfigure}{0.48\textwidth}
	    \includegraphics[width=3.4in]{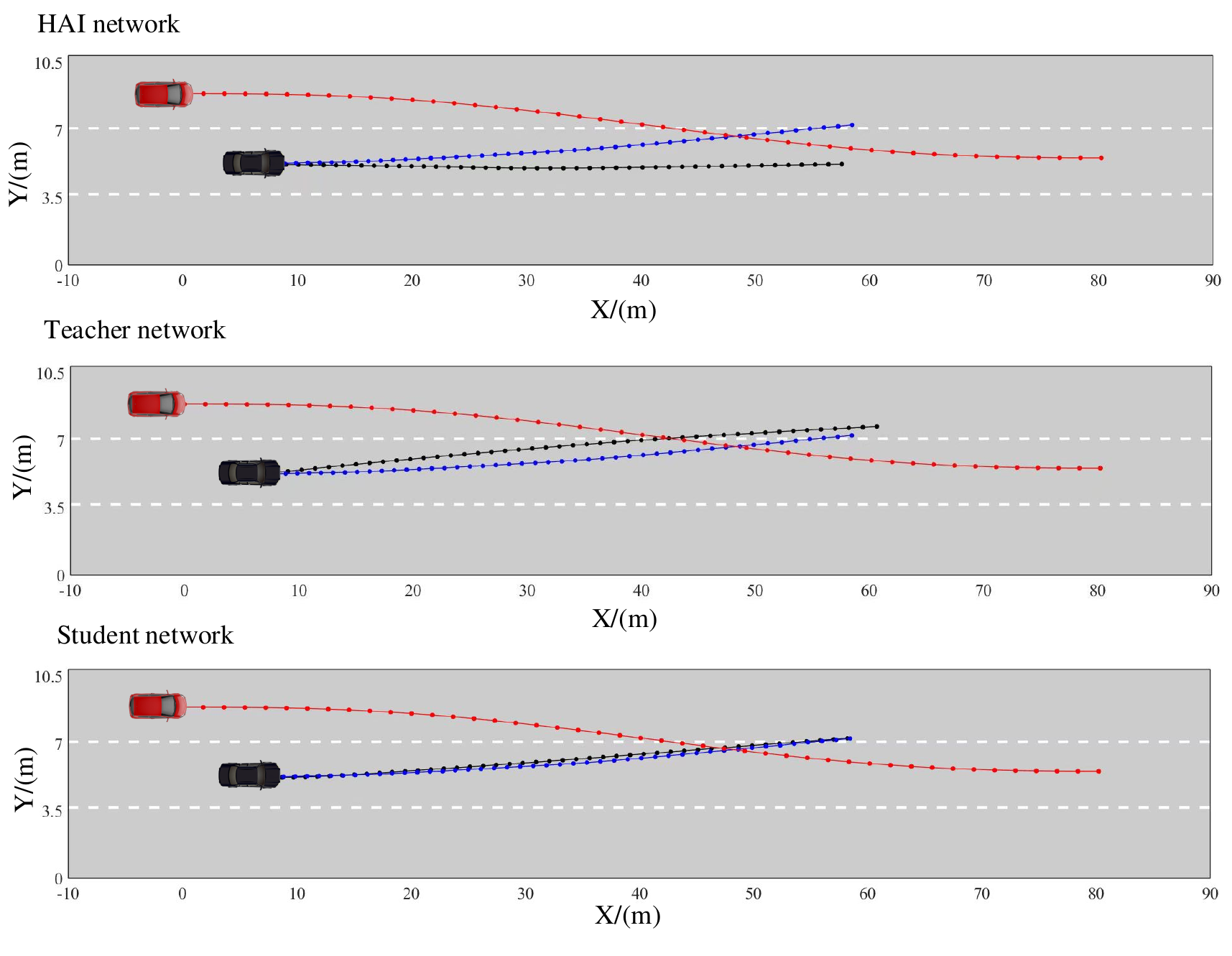}
	\caption{Driver Simulator Dataset.}
 \label{fig 19: The trajectory prediction results on the simulator test data in a high-speed driving scenario}
	\end{subfigure}
	\caption{The trajectory prediction results in a high-speed driving scenario.}
 \label{fig: prediction high-speed driving scenario}
\end{figure}%

To visually present the prediction results of different networks, we perform two V2V driving scenarios on both datasets: (i) a high-speed driving scenario and (ii) a low-speed driving scenario. The related results are shown in Fig. \ref{fig 12: The trajectory prediction results on the Waymo Open Dataset in a low-speed driving scenario}, Fig. \ref{fig 13: The trajectory prediction results on the Waymo Open Dataset in a high-speed driving scenario}, Figs. \ref{fig 18: The trajectory prediction results on the simulator test data in a low-speed driving scenario}, and \ref{fig 19: The trajectory prediction results on the simulator test data in a high-speed driving scenario} respectively. The red line represents the AV's trajectories during the V2V interaction scenarios, while the blue line and the black line indicate the HDV's real trajectory and predicted trajectory, respectively. It is clear the proposed predictor can achieve better trajectory prediction results. Furthermore, we can observe that the prediction accuracy slightly decreases in the high-speed driving scenario compared to the low-speed driving scenario. This is because AVs mainly operate at lower speeds during lane-change maneuvers, resulting in limited training data for high-speed driving scenarios.

\begin{figure}[ht]
	\centering
 \begin{subfigure}{0.48\textwidth}
 \centering
	\includegraphics[width=3.5in]{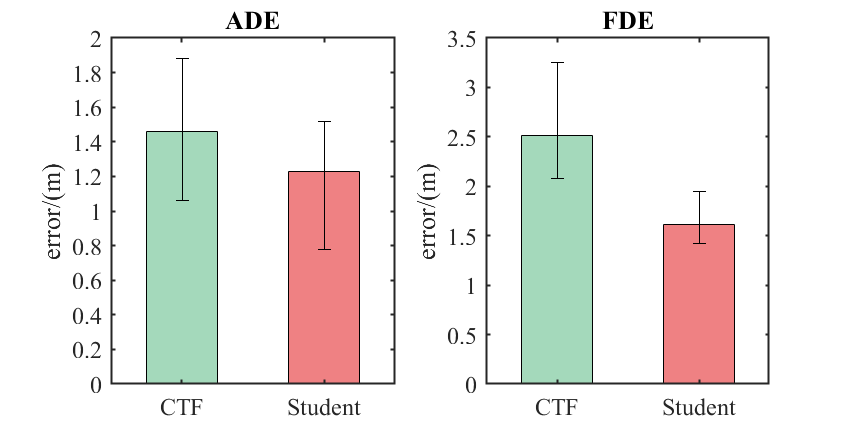}
	\caption{Waymo Open Dataset.}
 \label{fig 11:Comparison of trajectory prediction accuracy with different transfer learning methods on the Waymo Open Dataset}
 \end{subfigure}
	\begin{subfigure}{0.48\textwidth}
	\centering
	\includegraphics[width=3.5in]{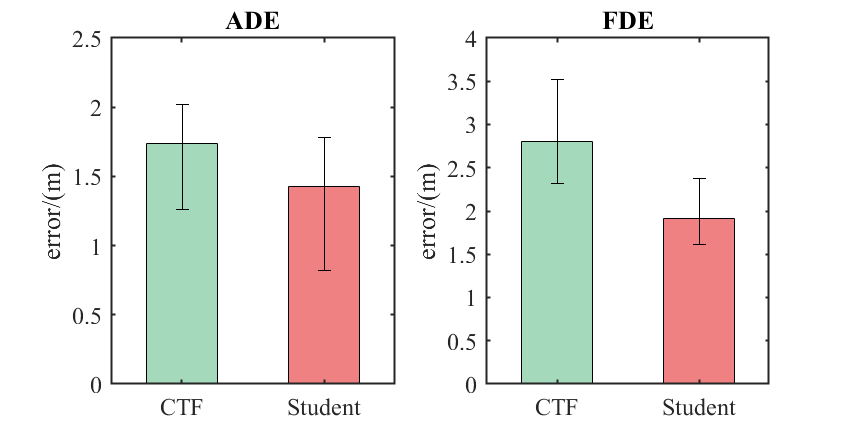}
	\caption{Vehicle Simulator Dataset.}
 \label{fig 17:Comparison of trajectory prediction accuracy with different transfer learning methods on the simulator test data}
  \end{subfigure}
\caption{Comparison of trajectory prediction accuracy with different transfer learning methods.}
\label{fig:different transfer learning methods}
\end{figure}%

\vspace{.2em} \noindent \textbf{Value of using hint-based transfer learning method}. To assess the performance of the proposed transfer-learning framework during the training process, we compare prediction results between the proposed method (i.e., student network trained with a hint-based transfer learning method) and a classic transfer-learning training framework (CTF) that employs only the knowledge distillation method. As shown in Fig. \ref{fig 11:Comparison of trajectory prediction accuracy with different transfer learning methods on the Waymo Open Dataset} and Fig. \ref{fig 17:Comparison of trajectory prediction accuracy with different transfer learning methods on the simulator test data}, it is clear that the proposed transfer learning method can significantly improve prediction accuracy by combining the hint-based training process and the classical knowledge distillation method. The indicators of ADE and FDE are reduced by $\ 15.95\%$ and $\ 36.80\%$, respectively, for the Waymo dataset, and by $\ 17.92\%$ and $\ 29.47\%$, respectively, for the driver simulator dataset. 

\begin{figure}[ht]
\centering
\begin{subfigure}{0.48\textwidth}
	\centering
	\includegraphics[width=3 in]{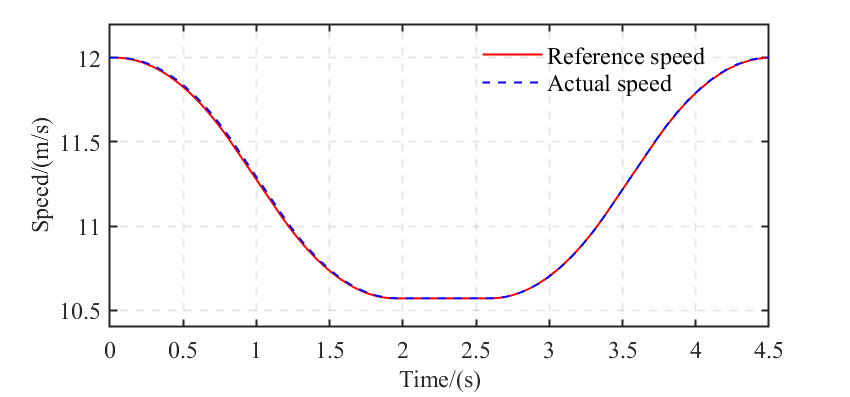}
	\caption{The speed profile.}
 \label{fig 14:The speed profile in test 1}
 \end{subfigure}

\begin{subfigure}{0.48\textwidth}
	\centering
	\includegraphics[width=3.2 in]{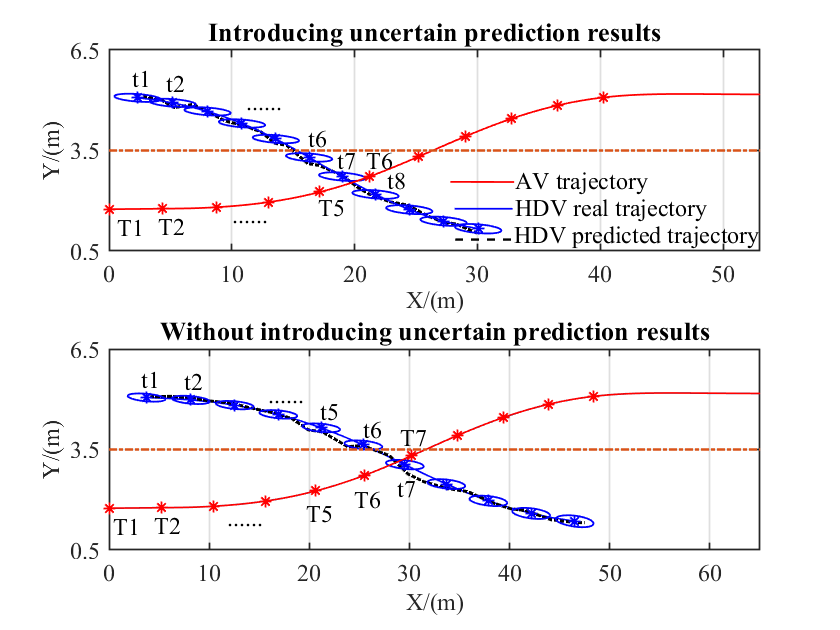}
	\caption{The trajectories of the AV and the HDV in test 1.}
 \label{fig 15:The trajectories of the AV and the HDV in test 1}
 \end{subfigure}

\begin{subfigure}{0.48\textwidth}
	\centering
	\includegraphics[width= 3 in]{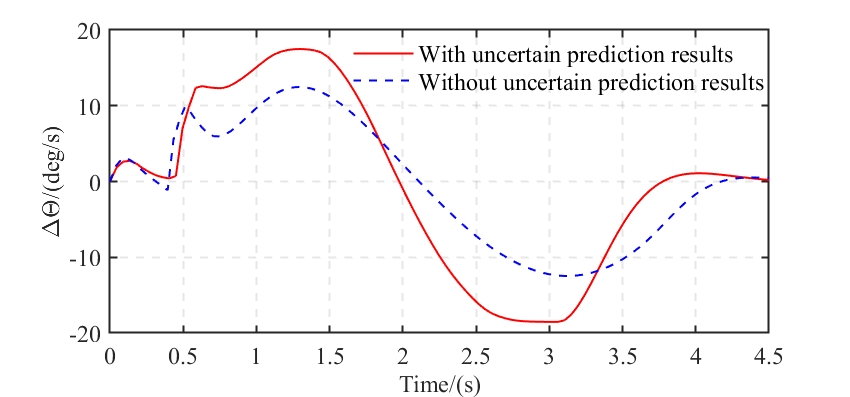}
	\caption{The change rate of the heading angle.}
 \label{fig 16:The change rate of the heading angle in test 1}
\end{subfigure}

\begin{subfigure}{0.48\textwidth}
	\centering
	\includegraphics[width= 3 in]{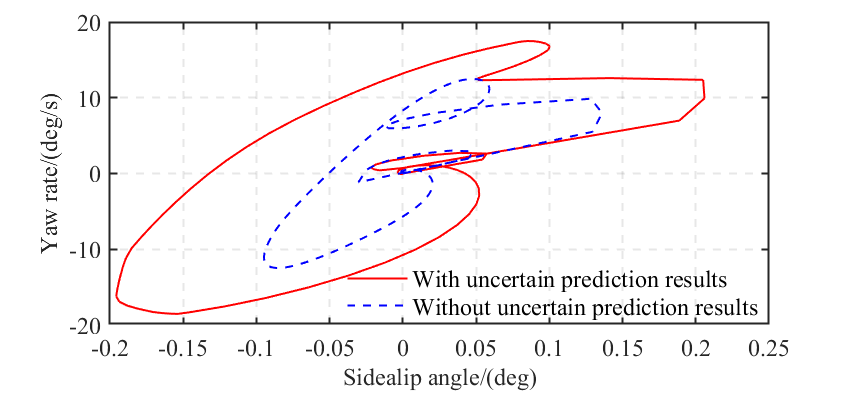}
	\caption{The phase plane of AV's lateral motion.}
 \label{fig 17:The phase plane of AV's lateral motion in test 1}
\end{subfigure}%
\caption{Control performance in Test 1.}
\label{fig:test 1}
\end{figure}%

\subsection{Verification of Introducing Uncertain Prediction Results into the Path Planning Process}

\begin{figure}[ht]
	\centering
	\includegraphics[width=3.3 in]{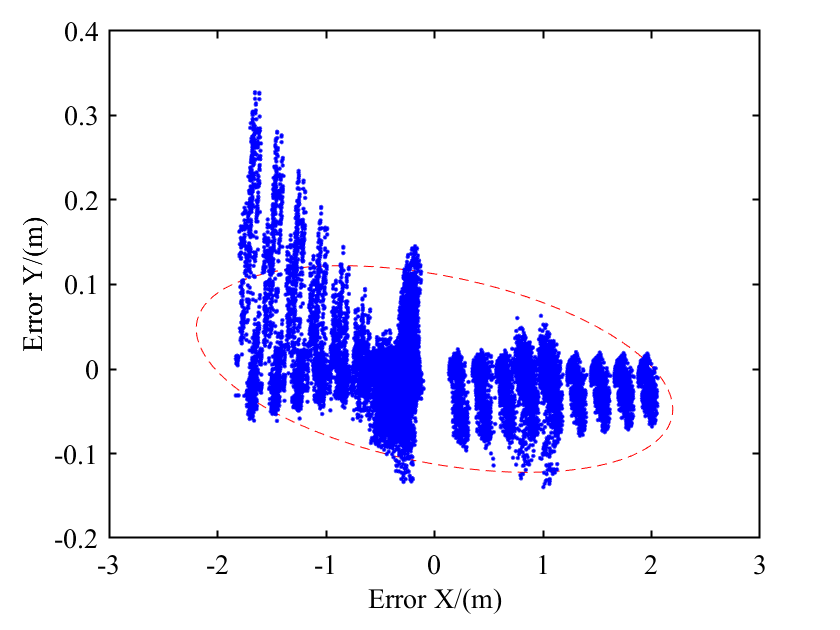}
	\caption{The uncertain representation of the predictor on the simulator test data.} \label{Fig20:The uncertain representation of the predictor on the simulator test data}
\end{figure}

We verify the effectiveness of introducing uncertain prediction results, i.e., inequality \ref{eq24: Safety distance}, into the path planning process.  Since the verification requires simulating the real-time interactions between the AV and HDV, we perform driver simulator-based experiments on V2V interaction scenarios with low-speed and high-speed driving conditions. We compare the motion control performance of the AV, including safety, stability, and comfort, with and without the incorporation of uncertainty quantification results into the path planning design. Here, we first use the the driver simulator data to construct an error eclipse of the proposed trajectory predictor, as shown in Fig.~\ref{Fig20:The uncertain representation of the predictor on the simulator test data}.

\vspace{0.5em} 
\noindent \emph{1) Low-speed Driving Scenario}
\vspace{0.5em}

The test results in the low-speed driving scenario are demonstrated in Fig. \ref{fig:test 1}. It is clear from the speed profile shown in Fig. \ref{fig 14:The speed profile in test 1} that the proposed motion control can guarantee satisfactory longitudinal tracking performance. Fig. \ref{fig 15:The trajectories of the AV and the HDV in test 1} illustrates the trajectories of the AV and HDV during a V2V interaction scenario involving lane-exchange maneuvers, including the predicted results from the trajectory predictor. The results show that the trajectory predictor's error consistently remains within the error ellipse shown in Fig. \ref{Fig20:The uncertain representation of the predictor on the simulator test data}. This demonstrates the effectiveness of the proposed uncertainty quantification method in describing the predictor's errors. We label the error ellipse along the predicted trajectories in the tests, as depicted in Fig. \ref{Fig20:The uncertain representation of the predictor on the simulator test data}. When the AV's trajectory passes through the error ellipse, it indicates a potential collision between vehicles. Furthermore, to better illustrate the results, we label the trajectory of the AV every three sampling instants as $T_i$, while the trajectory of the HDV every three sampling instants is labeled as $t_i$.

The path-planning ability with and without incorporating uncertain prediction results are compared. From the results, the AV can maintain a safe distance from the HDV when uncertain prediction results are introduced. For comparison, when uncertain prediction results are not introduced, the AV's trajectory approaches the error ellipse at the 7th sampling instant. However, as shown in inequality (\ref{eq24: Safety distance}), in addition to the constraints of the error ellipse, a safety threshold $S_{saf}$ must also be satisfied to guarantee a safe distance between vehicles. Hence, in this case, this trajectory cannot satisfy the safety constraint. Since uncertain prediction results are not introduced as a constraint, the TOPSIS strategy would select a trajectory with better performance indicators as proposed in section IV-C. This can be demonstrated in Figs. \ref{fig 16:The change rate of the heading angle in test 1} and \ref{fig 17:The phase plane of AV's lateral motion in test 1}. In terms of driving comfort shown in Fig. \ref{fig 16:The change rate of the heading angle in test 1}, due to a longer longitudinal lane-change distance as depicted in Fig. \ref{fig 15:The trajectories of the AV and the HDV in test 1}, the AV exhibits a smaller change rate of heading angle when uncertain prediction results are not considered, which represents better driving comfort. Furthermore, we introduce the classical "yaw rate-sideslip angle" phase plane \cite{liang2021distributed} to describe the AV's stability performance. As illustrated in Fig. \ref{fig 17:The phase plane of AV's lateral motion in test 1}, the AV operates within a smaller region in the phase plane when uncertain prediction results are not considered, indicating a more stable state. However, while relaxing the safety constraints of the AV may improve performance, it may also lead to potential collisions. This proves the effectiveness of introducing uncertain prediction results into the path-planning module to enhance the AV's safety.

\vspace{0.5em} 
\noindent \emph{2) High-speed Driving Scenario}
\vspace{0.5em}

\begin{figure}[ht]
\begin{subfigure}{0.48\textwidth}
	\centering
	\includegraphics[width=3in]{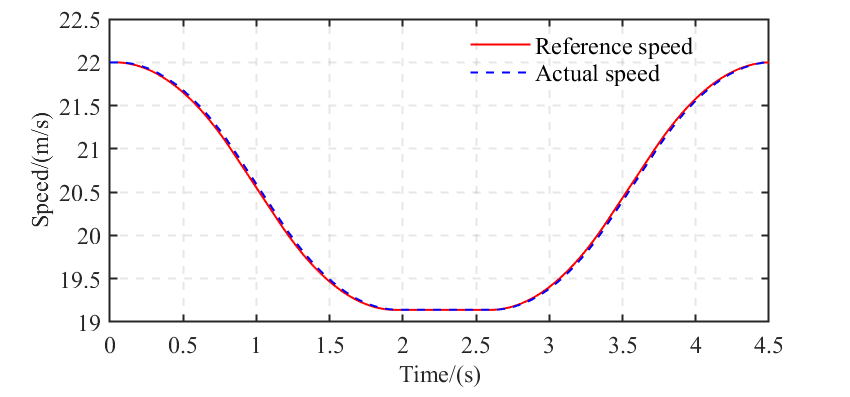}
	\caption{The speed profile.}
 \label{fig 18:The speed profile in test 2}
 \end{subfigure}

\begin{subfigure}{0.48\textwidth}
	\centering
	\includegraphics[width=3.2 in]{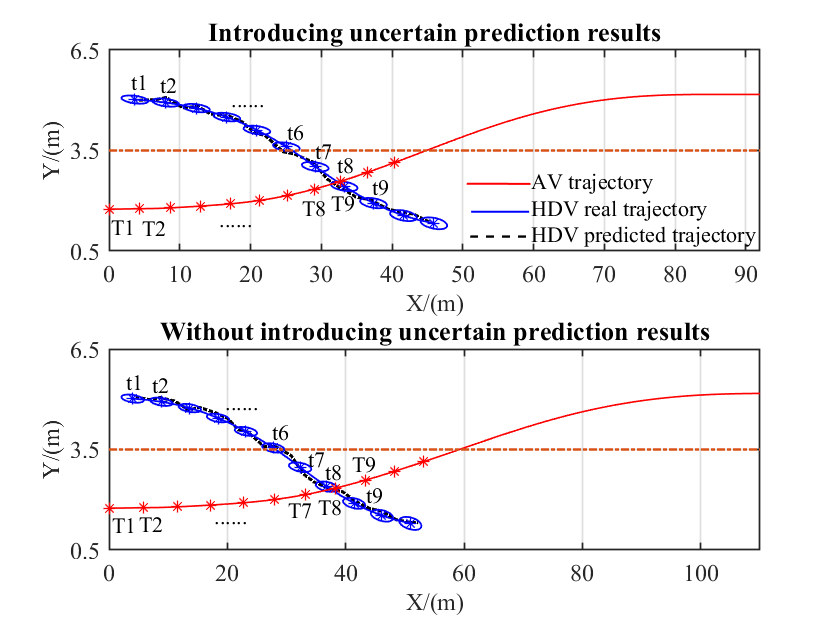}
	\caption{The trajectories of the AV and the HDV.}
 \label{fig 19:The trajectories of the AV and the HDV in test 2}
 \end{subfigure}

\begin{subfigure}{0.48\textwidth}
	\centering
	\includegraphics[width=3 in]{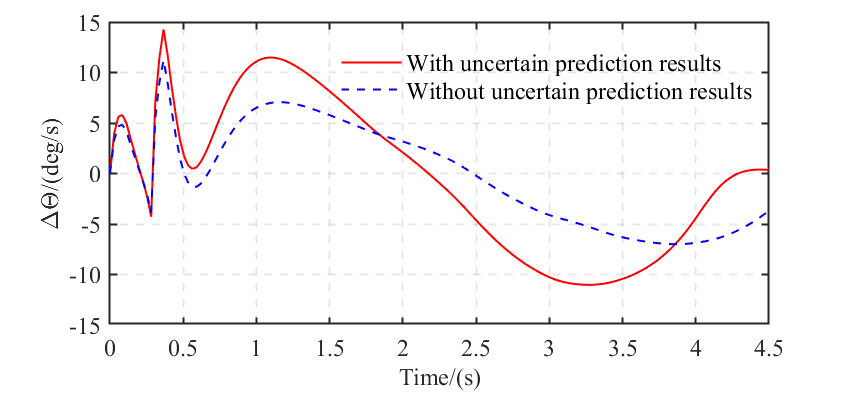}
	\caption{The change rate of the heading angle.}
 \label{fig 20:The change rate of the heading angle in test 2}
 \end{subfigure}

\begin{subfigure}{0.48\textwidth}
	\centering
	\includegraphics[width=3 in]{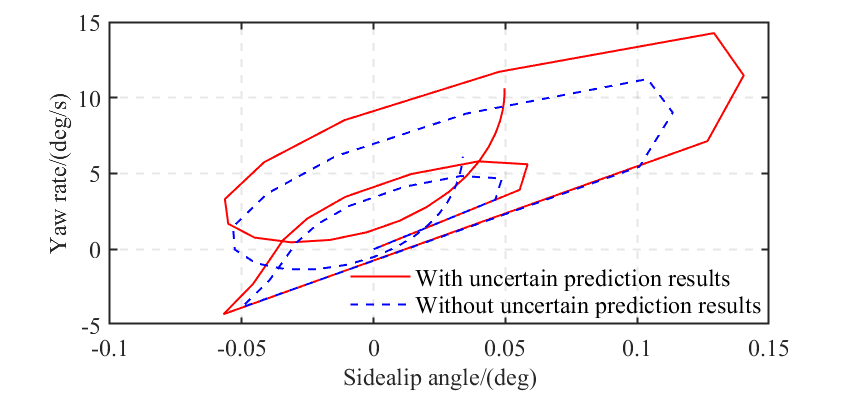}
	\caption{The phase plane of AV's lateral motion.}
 \label{fig 21:The phase plane of AV's lateral motion in test 2}
  \end{subfigure}
  \caption{Control performance in test 2.}
 \label{fig: test 2}
\end{figure}

Another test case under a high-speed driving condition is also conducted. Fig. \ref{fig 18:The speed profile in test 2} demonstrates the satisfactory longitudinal-motion tracking performance. From the comparative tests of trajectory profiles in Fig. \ref{fig 19:The trajectories of the AV and the HDV in test 2}, the error ellipse can accurately describe the uncertainty in prediction results. Furthermore, it is clear that the AV directly passes through the error ellipse at the 8th sampling instant when uncertain prediction results are not considered. This means that a potential collision could occur between vehicles when completing the lane-exchange maneuvers at high-speed driving conditions. The test results in Figs. \ref{fig 20:The change rate of the heading angle in test 2} and \ref{fig 21:The phase plane of AV's lateral motion in test 2} also demonstrate that, when disregarding the constraint of vehicle safety distance, the proposed TOPSIS strategy would generate a reference path that is significantly more stable and comfortable. These findings are consistent with the test results in the low-speed driving scenario. The test results in both low-speed and high-speed driving scenarios demonstrate the effectiveness of the proposed path planning in ensuring vehicle safety during V2V interaction scenarios involving lane-exchange maneuvers.

\section{CONCLUSION}
In this work, we propose a systematic framework for motion planning in a V2V interaction scenario, informed by an interaction-aware trajectory prediction algorithm. We further improve the performance of the conventional interaction-aware predictor by proposing an innovative transform-transfer learning-based training method. Specifically, we construct a transfer learning framework to transfer knowledge from a teacher network (i.e., the conventional trajectory predictor) to the student framework, utilizing limited interaction data in autonomous driving scenarios. Subsequently, we quantify the uncertainties in the prediction results characterized as an error ellipse and introduced into the path-planning process to impose safety constraints when evaluating the performance of different candidate paths.

The test results demonstrate the proposed framework can be feasible for motion planning in a V2V interaction scenario involving lane-exchange maneuvers. Compared with conventional methods (i.e., the teacher network), the proposed trajectory predictor can improve the performance indicators of ADE and FDE by $\ 34.97\%$ and $\ 57.45\%$, respectively, in the simulator test data. Considering the uncertainty of drivers’ behaviors in multi-agent scenarios, the future work can integrate more traffic information into the interaction-aware predictor to improve the performance, such as the driver’s states.



\bibliographystyle{IEEEtran}
\bibliography{ref1.bib}

\begin{thebibliography}{10}
\providecommand{\url}[1]{#1}
\csname url@samestyle\endcsname
\providecommand{\newblock}{\relax}
\providecommand{\bibinfo}[2]{#2}
\providecommand{\BIBentrySTDinterwordspacing}{\spaceskip=0pt\relax}
\providecommand{\BIBentryALTinterwordstretchfactor}{4}
\providecommand{\BIBentryALTinterwordspacing}{\spaceskip=\fontdimen2\font plus
\BIBentryALTinterwordstretchfactor\fontdimen3\font minus \fontdimen4\font\relax}
\providecommand{\BIBforeignlanguage}[2]{{%
\expandafter\ifx\csname l@#1\endcsname\relax
\typeout{** WARNING: IEEEtran.bst: No hyphenation pattern has been}%
\typeout{** loaded for the language `#1'. Using the pattern for}%
\typeout{** the default language instead.}%
\else
\language=\csname l@#1\endcsname
\fi
#2}}
\providecommand{\BIBdecl}{\relax}
\BIBdecl

\bibitem{petrovic2020traffic}
{\DJ}.~Petrovi{\'c}, R.~Mijailovi{\'c}, and D.~Pe{\v{s}}i{\'c}, ``Traffic accidents with autonomous vehicles: type of collisions, manoeuvres and errors of conventional vehicles’ drivers,'' \emph{Transportation research procedia}, vol.~45, pp. 161--168, 2020.

\bibitem{liang2022mas}
J.~Liang, Y.~Li, G.~Yin, L.~Xu, Y.~Lu, J.~Feng, T.~Shen, and G.~Cai, ``A mas-based hierarchical architecture for the cooperation control of connected and automated vehicles,'' \emph{IEEE Transactions on Vehicular Technology}, vol.~72, no.~2, pp. 1559--1573, 2022.

\bibitem{liang2023polytopic}
J.~Liang, Q.~Tian, J.~Feng, D.~Pi, and G.~Yin, ``A polytopic model-based robust predictive control scheme for path tracking of autonomous vehicles,'' \emph{IEEE Transactions on Intelligent Vehicles}, 2023.

\bibitem{yue2023cooperative}
W.~Yue, C.~Li, S.~Wang, N.~Xue, and J.~Wu, ``Cooperative incident management in mixed traffic of cavs and human-driven vehicles,'' \emph{IEEE Transactions on Intelligent Transportation Systems}, 2023.

\bibitem{sun2020cooperative}
Z.~Sun, T.~Huang, and P.~Zhang, ``Cooperative decision-making for mixed traffic: A ramp merging example,'' \emph{Transportation research part C: emerging technologies}, vol. 120, p. 102764, 2020.

\bibitem{zhou2024enhancing}
J.~Zhou, L.~Yan, and K.~Yang, ``Enhancing system-level safety in mixed-autonomy platoon via safe reinforcement learning,'' \emph{IEEE Transactions on Intelligent Vehicles}, 2024.

\bibitem{zhou2024parameter}
J.~Zhou and K.~Yang, ``A parameter privacy-preserving strategy for mixed-autonomy platoon control,'' \emph{arXiv preprint arXiv:2401.15561}, 2024.

\bibitem{shawky2020factors}
M.~Shawky, ``Factors affecting lane change crashes,'' \emph{IATSS research}, vol.~44, no.~2, pp. 155--161, 2020.

\bibitem{singh2021analyzing}
H.~Singh and A.~Kathuria, ``Analyzing driver behavior under naturalistic driving conditions: A review,'' \emph{Accident Analysis \& Prevention}, vol. 150, p. 105908, 2021.

\bibitem{mozaffari2020deep}
S.~Mozaffari, O.~Y. Al-Jarrah, M.~Dianati, P.~Jennings, and A.~Mouzakitis, ``Deep learning-based vehicle behavior prediction for autonomous driving applications: A review,'' \emph{IEEE Transactions on Intelligent Transportation Systems}, vol.~23, no.~1, pp. 33--47, 2020.

\bibitem{xu2022aggressive}
W.~Xu, J.~Wang, T.~Fu, H.~Gong, and A.~Sobhani, ``Aggressive driving behavior prediction considering driver’s intention based on multivariate-temporal feature data,'' \emph{Accident Analysis \& Prevention}, vol. 164, p. 106477, 2022.

\bibitem{liao2023driver}
X.~Liao, X.~Zhao, Z.~Wang, Z.~Zhao, K.~Han, R.~Gupta, M.~J. Barth, and G.~Wu, ``Driver digital twin for online prediction of personalized lane change behavior,'' \emph{IEEE Internet of Things Journal}, 2023.

\bibitem{wang2021intelligent}
W.~Wang, T.~Qie, C.~Yang, W.~Liu, C.~Xiang, and K.~Huang, ``An intelligent lane-changing behavior prediction and decision-making strategy for an autonomous vehicle,'' \emph{IEEE transactions on industrial electronics}, vol.~69, no.~3, pp. 2927--2937, 2021.

\bibitem{lefevre2014survey}
S.~Lef{\`e}vre, D.~Vasquez, and C.~Laugier, ``A survey on motion prediction and risk assessment for intelligent vehicles,'' \emph{ROBOMECH journal}, vol.~1, pp. 1--14, 2014.

\bibitem{gao2021trajectory}
H.~Gao, H.~Su, Y.~Cai, R.~Wu, Z.~Hao, Y.~Xu, W.~Wu, J.~Wang, Z.~Li, and Z.~Kan, ``Trajectory prediction of cyclist based on dynamic bayesian network and long short-term memory model at unsignalized intersections,'' \emph{Science China Information Sciences}, vol.~64, no.~7, p. 172207, 2021.

\bibitem{liu2020early}
Q.~Liu, S.~Xu, C.~Lu, H.~Yao, and H.~Chen, ``Early recognition of driving intention for lane change based on recurrent hidden semi-markov model,'' \emph{IEEE transactions on vehicular technology}, vol.~69, no.~10, pp. 10\,545--10\,557, 2020.

\bibitem{girma2020deep}
A.~Girma, S.~Amsalu, A.~Workineh, M.~Khan, and A.~Homaifar, ``Deep learning with attention mechanism for predicting driver intention at intersection,'' in \emph{2020 IEEE Intelligent Vehicles Symposium (IV)}.\hskip 1em plus 0.5em minus 0.4em\relax IEEE, 2020, pp. 1183--1188.

\bibitem{phillips2017generalizable}
D.~J. Phillips, T.~A. Wheeler, and M.~J. Kochenderfer, ``Generalizable intention prediction of human drivers at intersections,'' in \emph{2017 IEEE intelligent vehicles symposium (IV)}.\hskip 1em plus 0.5em minus 0.4em\relax IEEE, 2017, pp. 1665--1670.

\bibitem{mcnaughton2011motion}
M.~McNaughton, C.~Urmson, J.~M. Dolan, and J.-W. Lee, ``Motion planning for autonomous driving with a conformal spatiotemporal lattice,'' in \emph{2011 IEEE International Conference on Robotics and Automation}.\hskip 1em plus 0.5em minus 0.4em\relax IEEE, 2011, pp. 4889--4895.

\bibitem{schwarting2018planning}
W.~Schwarting, J.~Alonso-Mora, and D.~Rus, ``Planning and decision-making for autonomous vehicles,'' \emph{Annual Review of Control, Robotics, and Autonomous Systems}, vol.~1, pp. 187--210, 2018.

\bibitem{alahi2016social}
A.~Alahi, K.~Goel, V.~Ramanathan, A.~Robicquet, L.~Fei-Fei, and S.~Savarese, ``Social lstm: Human trajectory prediction in crowded spaces,'' in \emph{Proceedings of the IEEE conference on computer vision and pattern recognition}, 2016, pp. 961--971.

\bibitem{deo2018convolutional}
N.~Deo and M.~M. Trivedi, ``Convolutional social pooling for vehicle trajectory prediction,'' in \emph{Proceedings of the IEEE conference on computer vision and pattern recognition workshops}, 2018, pp. 1468--1476.

\bibitem{song2020pip}
H.~Song, W.~Ding, Y.~Chen, S.~Shen, M.~Y. Wang, and Q.~Chen, ``Pip: Planning-informed trajectory prediction for autonomous driving,'' in \emph{Computer Vision--ECCV 2020: 16th European Conference, Glasgow, UK, August 23--28, 2020, Proceedings, Part XXI 16}.\hskip 1em plus 0.5em minus 0.4em\relax Springer, 2020, pp. 598--614.

\bibitem{chen2022efficient}
Y.~Chen, R.~Xin, J.~Cheng, Q.~Zhang, X.~Mei, M.~Liu, and L.~Wang, ``Efficient speed planning for autonomous driving in dynamic environment with interaction point model,'' \emph{IEEE Robotics and Automation Letters}, vol.~7, no.~4, pp. 11\,839--11\,846, 2022.

\bibitem{espinoza2022deep}
J.~L.~V. Espinoza, A.~Liniger, W.~Schwarting, D.~Rus, and L.~Van~Gool, ``Deep interactive motion prediction and planning: Playing games with motion prediction models,'' in \emph{Learning for Dynamics and Control Conference}.\hskip 1em plus 0.5em minus 0.4em\relax PMLR, 2022, pp. 1006--1019.

\bibitem{ngiam2021scene}
J.~Ngiam, B.~Caine, V.~Vasudevan, Z.~Zhang, H.-T.~L. Chiang, J.~Ling, R.~Roelofs, A.~Bewley, C.~Liu, A.~Venugopal \emph{et~al.}, ``Scene transformer: A unified architecture for predicting multiple agent trajectories,'' \emph{arXiv preprint arXiv:2106.08417}, 2021.

\bibitem{zhang2022trajectory}
K.~Zhang, X.~Feng, L.~Wu, and Z.~He, ``Trajectory prediction for autonomous driving using spatial-temporal graph attention transformer,'' \emph{IEEE Transactions on Intelligent Transportation Systems}, vol.~23, no.~11, pp. 22\,343--22\,353, 2022.

\bibitem{sadeghian2019sophie}
A.~Sadeghian, V.~Kosaraju, A.~Sadeghian, N.~Hirose, H.~Rezatofighi, and S.~Savarese, ``Sophie: An attentive gan for predicting paths compliant to social and physical constraints,'' in \emph{Proceedings of the IEEE/CVF conference on computer vision and pattern recognition}, 2019, pp. 1349--1358.

\bibitem{gupta2018social}
A.~Gupta, J.~Johnson, L.~Fei-Fei, S.~Savarese, and A.~Alahi, ``Social gan: Socially acceptable trajectories with generative adversarial networks,'' in \emph{Proceedings of the IEEE conference on computer vision and pattern recognition}, 2018, pp. 2255--2264.

\bibitem{kingma2013auto}
D.~P. Kingma and M.~Welling, ``Auto-encoding variational bayes,'' \emph{arXiv preprint arXiv:1312.6114}, 2013.

\bibitem{lee2017desire}
N.~Lee, W.~Choi, P.~Vernaza, C.~B. Choy, P.~H. Torr, and M.~Chandraker, ``Desire: Distant future prediction in dynamic scenes with interacting agents,'' in \emph{Proceedings of the IEEE conference on computer vision and pattern recognition}, 2017, pp. 336--345.

\bibitem{wen2023analysis}
X.~Wen, C.~Huang, S.~Jian, and D.~He, ``Analysis of discretionary lane-changing behaviours of autonomous vehicles based on real-world data,'' \emph{Transportmetrica A: Transport Science}, pp. 1--24, 2023.

\bibitem{liang2021distributed}
J.~Liang, Y.~Lu, G.~Yin, Z.~Fang, W.~Zhuang, Y.~Ren, L.~Xu, and Y.~Li, ``A distributed integrated control architecture of afs and dyc based on mas for distributed drive electric vehicles,'' \emph{IEEE transactions on vehicular technology}, vol.~70, no.~6, pp. 5565--5577, 2021.

\bibitem{wang2023lane}
Z.~Wang, J.~Guo, Z.~Hu, H.~Zhang, J.~Zhang, and J.~Pu, ``Lane transformer: A high-efficiency trajectory prediction model,'' \emph{IEEE Open Journal of Intelligent Transportation Systems}, vol.~4, pp. 2--13, 2023.

\bibitem{chen2022vehicle}
X.~Chen, H.~Zhang, F.~Zhao, Y.~Cai, H.~Wang, and Q.~Ye, ``Vehicle trajectory prediction based on intention-aware non-autoregressive transformer with multi-attention learning for internet of vehicles,'' \emph{IEEE Transactions on Instrumentation and Measurement}, vol.~71, pp. 1--12, 2022.

\bibitem{veit2016residual}
A.~Veit, M.~J. Wilber, and S.~Belongie, ``Residual networks behave like ensembles of relatively shallow networks,'' \emph{Advances in neural information processing systems}, vol.~29, 2016.

\bibitem{huang2022survey}
Y.~Huang, J.~Du, Z.~Yang, Z.~Zhou, L.~Zhang, and H.~Chen, ``A survey on trajectory-prediction methods for autonomous driving,'' \emph{IEEE Transactions on Intelligent Vehicles}, vol.~7, no.~3, pp. 652--674, 2022.

\bibitem{bae2019layer}
J.-H. Bae, J.~Yim, N.-S. Kim, C.-S. Pyo, and J.~Kim, ``Layer-wise hint-based training for knowledge transfer in a teacher-student framework,'' \emph{ETRI Journal}, vol.~41, no.~2, pp. 242--253, 2019.

\bibitem{adriana2015fitnets}
R.~Adriana, B.~Nicolas, K.~S. Ebrahimi, C.~Antoine, G.~Carlo, and B.~Yoshua, ``Fitnets: Hints for thin deep nets,'' \emph{Proc. ICLR}, vol.~2, no.~3, p.~1, 2015.

\bibitem{hinton2015distilling}
G.~Hinton, O.~Vinyals, and J.~Dean, ``Distilling the knowledge in a neural network,'' \emph{arXiv preprint arXiv:1503.02531}, 2015.

\bibitem{chen2017learning}
G.~Chen, W.~Choi, X.~Yu, T.~Han, and M.~Chandraker, ``Learning efficient object detection models with knowledge distillation,'' \emph{Advances in neural information processing systems}, vol.~30, 2017.

\bibitem{girshick2015fast}
R.~Girshick, ``Fast r-cnn,'' in \emph{Proceedings of the IEEE international conference on computer vision}, 2015, pp. 1440--1448.

\bibitem{lancaster2005chi}
H.~O. Lancaster and E.~Seneta, ``Chi-square distribution,'' \emph{Encyclopedia of biostatistics}, vol.~2, 2005.

\bibitem{huang2021personalized}
C.~Huang, H.~Huang, P.~Hang, H.~Gao, J.~Wu, Z.~Huang, and C.~Lv, ``Personalized trajectory planning and control of lane-change maneuvers for autonomous driving,'' \emph{IEEE Transactions on Vehicular Technology}, vol.~70, no.~6, pp. 5511--5523, 2021.

\bibitem{li2023dynamic}
Y.~Li, L.~Li, D.~Ni \emph{et~al.}, ``Dynamic trajectory planning for automated lane changing using the quintic polynomial curve,'' \emph{Journal of Advanced Transportation}, vol. 2023, 2023.

\bibitem{martinez2017assessment}
J.~R.~G. Mart{\'\i}nez, J.~R. Res{\'e}ndiz, M.~{\'A}.~M. Prado, and E.~E.~C. Miguel, ``Assessment of jerk performance s-curve and trapezoidal velocity profiles,'' in \emph{2017 XIII international engineering congress (CONIIN)}.\hskip 1em plus 0.5em minus 0.4em\relax IEEE, 2017, pp. 1--7.

\bibitem{liang2023energy}
J.~Liang, J.~Feng, Z.~Fang, Y.~Lu, G.~Yin, X.~Mao, J.~Wu, and F.~Wang, ``An energy-oriented torque-vector control framework for distributed drive electric vehicles,'' \emph{IEEE Transactions on Transportation Electrification}, vol.~9, no.~3, pp. 4014--4031, 2023.

\end{thebibliography}

\vspace{11pt}
\begin{IEEEbiography}
[{\includegraphics[width=1in,height=1.25in,clip,keepaspectratio]{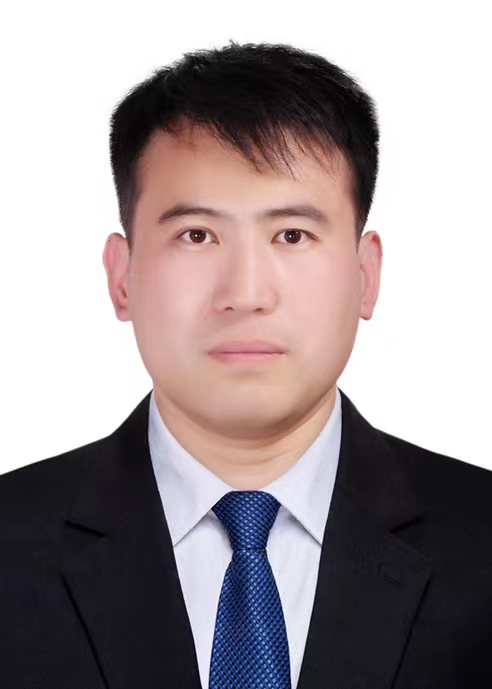}}]{Jinhao Liang}
received the B.S. degree from School of Mechanical Engineering, Nanjing University of Science and Technology, Nanjing, China, in 2017, and Ph.D. degree from School of Mechanical Engineering, Southeast University, Nanjing, China, in 2022. Now he is a Research Fellow with Department of Civil and Environmental Engineering, National University of Singapore. His research interests include vehicle dynamics and control, autonomous vehicles, and vehicle safety assistance system.
\end{IEEEbiography}

\vspace{11pt}
\begin{IEEEbiography}
[{\includegraphics[width=1in,height=1.25in,clip,keepaspectratio]{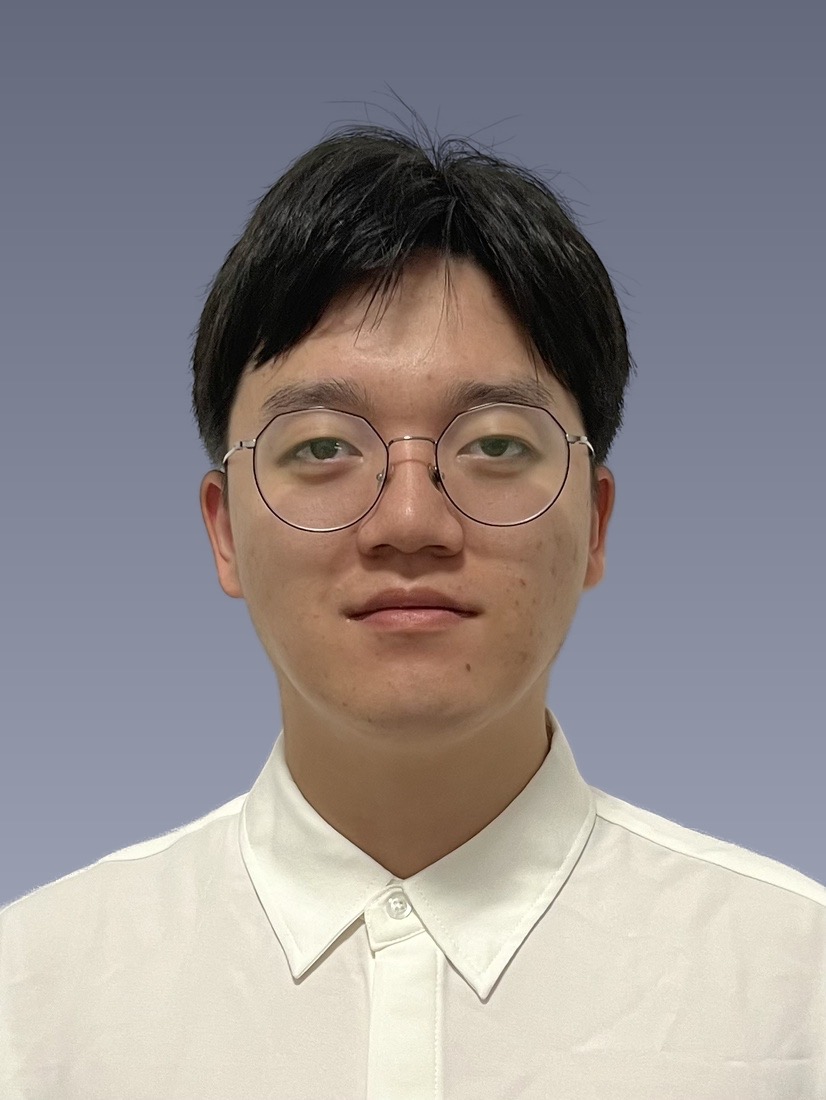}}]{Chaopeng Tan}
received his B.S. and Ph.D. degrees in traffic engineering from Tongji University, Shanghai, China in 2017 and 2022, respectively. He was a Postdoctoral Research Fellow with the Department of Civil and Environment Engineering, National University of Singapore from 2022 to 2024. He is currently a Postdoctoral Research Fellow with the Department of Transport and Planning, Delft University of Technology, The Netherlands. His main research interests include intelligent transportation systems, traffic modeling and control with connected vehicles, and privacy-preserving traffic control.
\end{IEEEbiography}

\vspace{11pt}
\begin{IEEEbiography}
[{\includegraphics[width=1in,height=1.25in,clip,keepaspectratio]{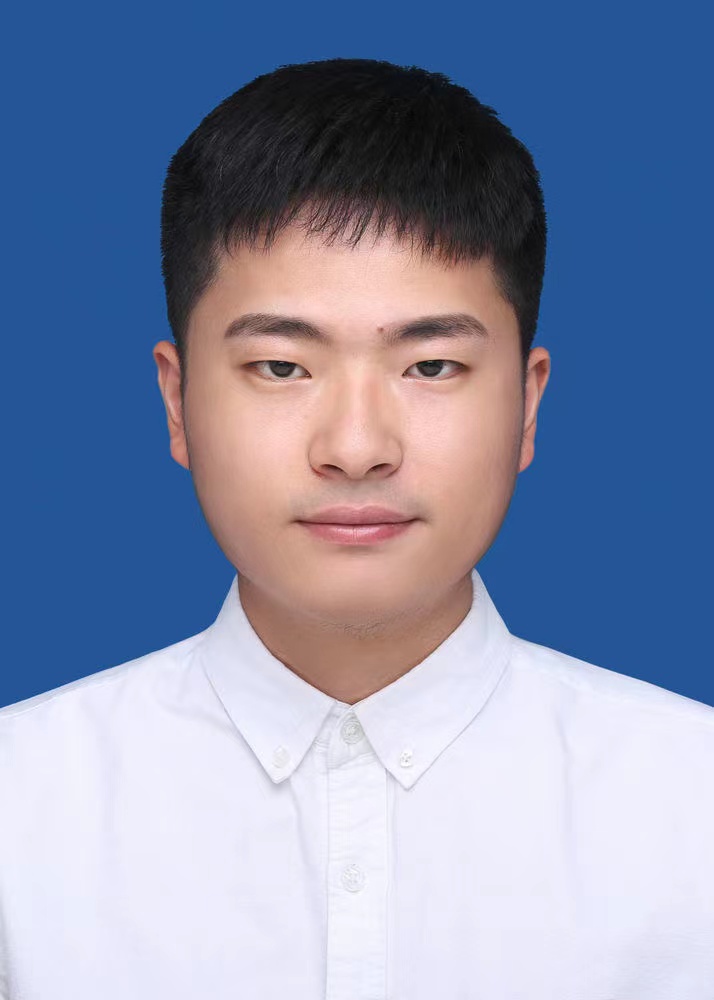}}]{Longhao Yan} received the B.Eng. degree and M.Eng. degree in School of Electronics and Control Engineering from Chang’an University, Xi’an, China, in 2019 and 2022 respectively. He is currently working towards a Ph.D. degree with the National University of Singapore. His research interests include lateral control and trajectory prediction of intelligent transportation systems.
\end{IEEEbiography}

\vspace{11pt}
\begin{IEEEbiography}
[{\includegraphics[width=1in,height=1.25in,clip,keepaspectratio]{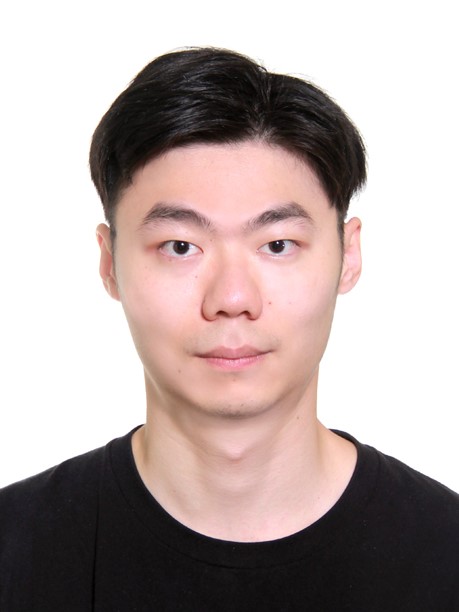}}]{Jingyuan Zhou} received the B.Eng. degree in Electronic Information Science and Technology from Sun Yat-sen University, Guangzhou, China, in 2022. He is currently working towards a Ph.D. degree with the National University of Singapore. His research interests include safe and secure connected and automated vehicles control in intelligent transportation systems.
\end{IEEEbiography}

\vspace{11pt}
\begin{IEEEbiography}
[{\includegraphics[width=1in,height=1.25in,clip,keepaspectratio]{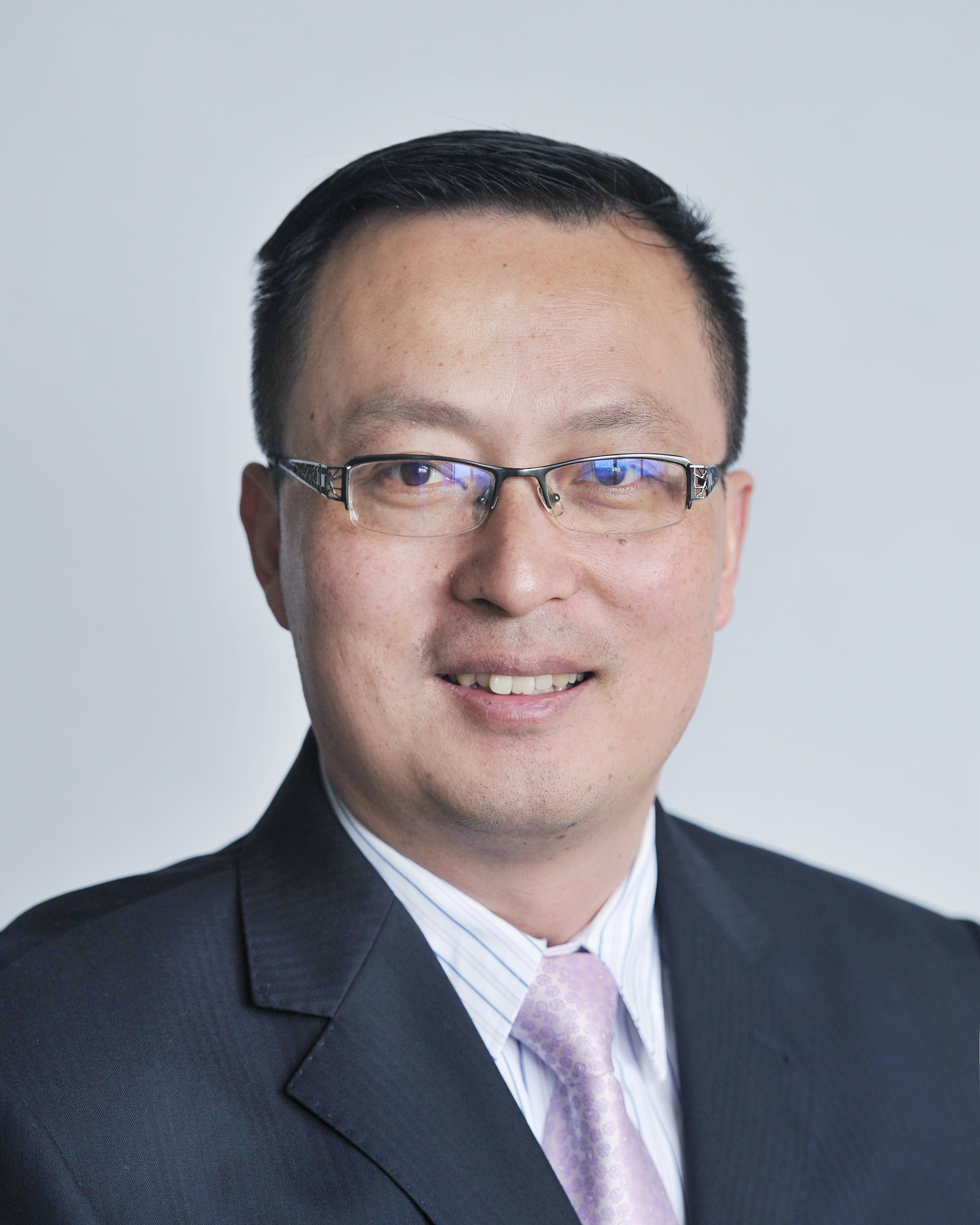}}]{Guodong Yin}
received the Ph.D. degree in mechanical engineering from Southeast University, Nanjing, China, in 2007. From August 2011 to August 2012, he was a Visiting Scholar with the Department of Mechanical and Aerospace Engineering, The Ohio State University, Columbus, OH, USA.
He is currently a Professor with the School of Mechanical Engineering, Southeast University. He was the recipient of the National Science Fund for Distinguished Young Scholars. His research interests include vehicle dynamics and control, automated vehicles, and connected vehicles.
\end{IEEEbiography}

\vspace{11pt}
\begin{IEEEbiography}
[{\includegraphics[width=1in,height=1.25in,clip,keepaspectratio]{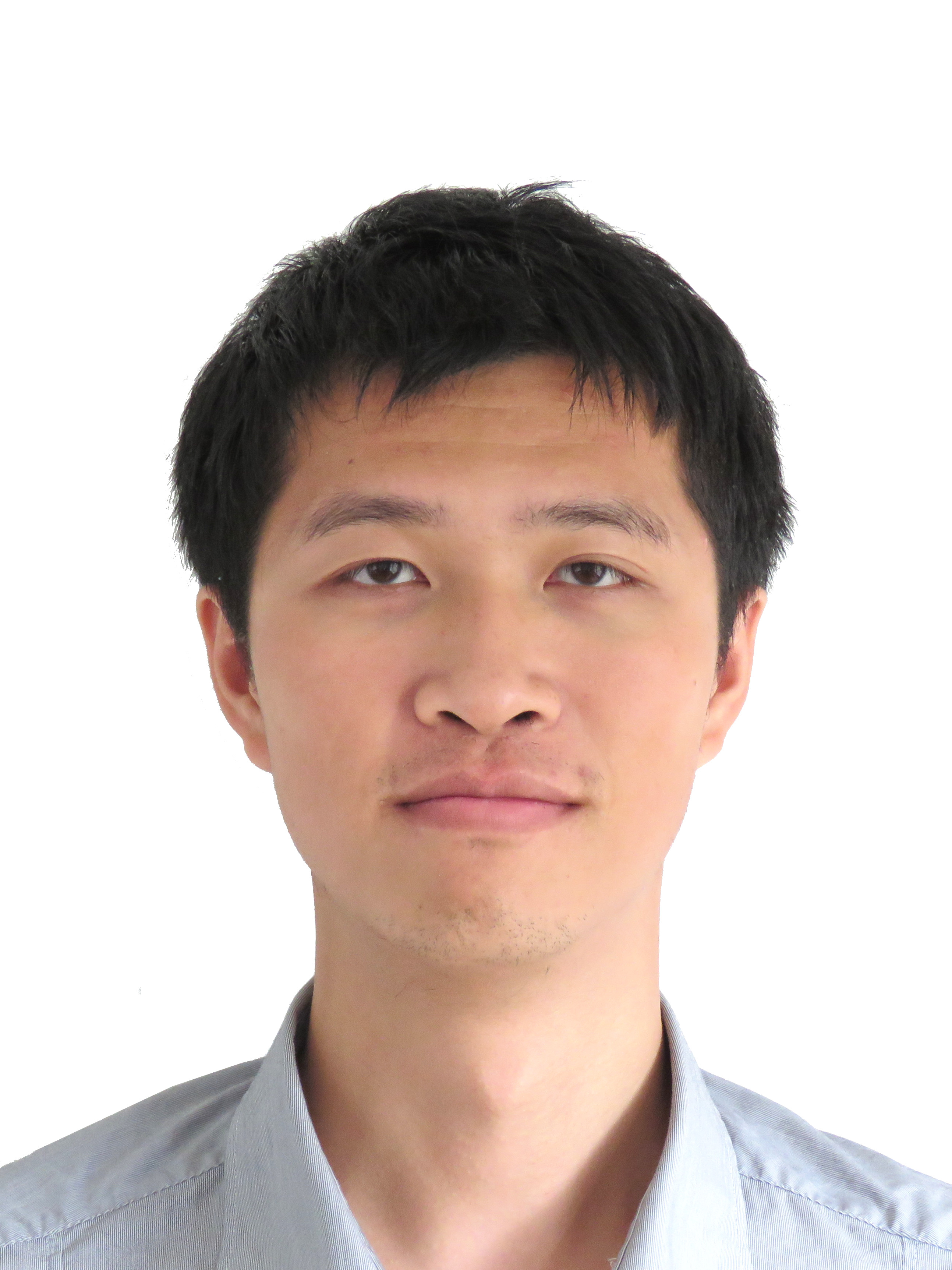}}]{Kaidi Yang}  is an Assistant Professor in the Department of Civil and Environmental Engineering at the National University of Singapore. Prior to this, he was a postdoctoral researcher with the Autonomous Systems Lab at Stanford University. He obtained a PhD degree from ETH Zurich and M.Sc. and B.Eng. degrees from Tsinghua University. His main research interest is the operation of future mobility systems enabled by connected and automated vehicles (CAVs) and shared mobility.
\end{IEEEbiography}
\vfill

\end{document}